\documentclass[lettersize,journal]{IEEEtran}
\usepackage{amsmath,amsfonts}
\usepackage{algorithmic}
\usepackage{algorithm}
\usepackage{array}
\usepackage[caption=false,font=normalsize,labelfont=sf,textfont=sf]{subfig}
\usepackage{bm}
\usepackage{textcomp}
\usepackage{stfloats}
\usepackage{url}
\usepackage{verbatim}
\usepackage{graphicx}
\usepackage{cite}
\usepackage{booktabs}
\usepackage{amssymb}
\usepackage{multirow}
\usepackage{color}
\usepackage{float}
\usepackage{latexsym}

\begin{document}

\title{A Transformer-based Model with Self-distillation for Multimodal Emotion Recognition in Conversations}

\author{Hui Ma, Jian Wang, Hongfei Lin, Bo Zhang, Yijia Zhang, and Bo Xu
\thanks{Hui Ma, Jian Wang (Corresponding author), Hongfei Lin, Bo Zhang and Bo Xu are with the School of Computer Science and Technology, Dalian University of Technology, Dalian 116024, China (e-mail:huima@mail.dlut.edu.cn; wangjian@dlut.edu.cn; hflin@dlut.edu.cn; zhangbo1998@mail.dlut.edu.cn; xubo@dlut.edu.cn).

Yijia Zhang is with the School of Information Science and Technology, Dalian Maritime
University, Dalian 116024, China (e-mail:zhangyijia@dlmu.edu.cn)}}




\maketitle

\begin{abstract}
Emotion recognition in conversations (ERC), the task of recognizing the emotion of each utterance in a conversation, is crucial for building empathetic machines. Existing studies focus mainly on capturing context- and speaker-sensitive dependencies on the textual modality but ignore the significance of multimodal information. Different from emotion recognition in textual conversations, capturing intra- and inter-modal interactions between utterances, learning weights between different modalities, and enhancing modal representations play important roles in multimodal ERC. In this paper, we propose a transformer-based model with self-distillation (SDT)\footnote{The code is available at \url{https://github.com/butterfliesss/SDT}.} for the task. The transformer-based model captures intra- and inter-modal interactions by utilizing intra- and inter-modal transformers, and learns weights between modalities dynamically by designing a hierarchical gated fusion strategy. Furthermore, to learn more expressive modal representations, we treat soft labels of the proposed model as extra training supervision. Specifically, we introduce self-distillation to transfer knowledge of hard and soft labels from the proposed model to each modality. Experiments on IEMOCAP and MELD datasets demonstrate that SDT outperforms previous state-of-the-art baselines.
\end{abstract}


\section{Introduction}
\IEEEPARstart{E}{motion} recognition in conversations (ERC) aims to automatically recognize the emotion of each utterance in a conversation. The task has recently become an important research topic due to its wide applications in opinion mining \cite{kumar:emotion2015}, health care \cite{pujo:emotion2019}, and building empathic dialogue systems \cite{zhou2020design}, etc. Unlike traditional emotion recognition (ER) on context-free sentences, modeling context- and speaker-sensitive dependencies lie at the heart of ERC.

Existing mainstream works on ERC can generally be categorized into sequence- and graph-based methods. Sequence-based methods \cite{hazarika:cmn2018, hazarika:icon2018, majumder:dialoguernn2019, jiao:higru2019, li:hitrans2020,ma2021han,mao:dialoguetrm2021,ma:mvn2022} use recurrent neural networks or transformers to model long-distance contextual information in a conversation. In contrast, graph-based methods \cite{ghosal:dialoguegcn2019,hu:mmgcn2021,nie-igcn,ren-lrgcn} design graph structures for conversations and then use graph neural networks to capture multiple dependencies. Although these methods show promising performance, most of them focus primarily on textual conversations without leveraging other modalities (i.e., acoustic and visual modalities). According to Mehrabian \cite{mehrabian1971silent}, people express emotions in a variety of ways, including verbal, vocal, and facial expressions. Therefore, multimodal information is more useful for understanding emotions than unimodal information.

\begin{figure}[!t]
\centering
\includegraphics[width=3.505in]{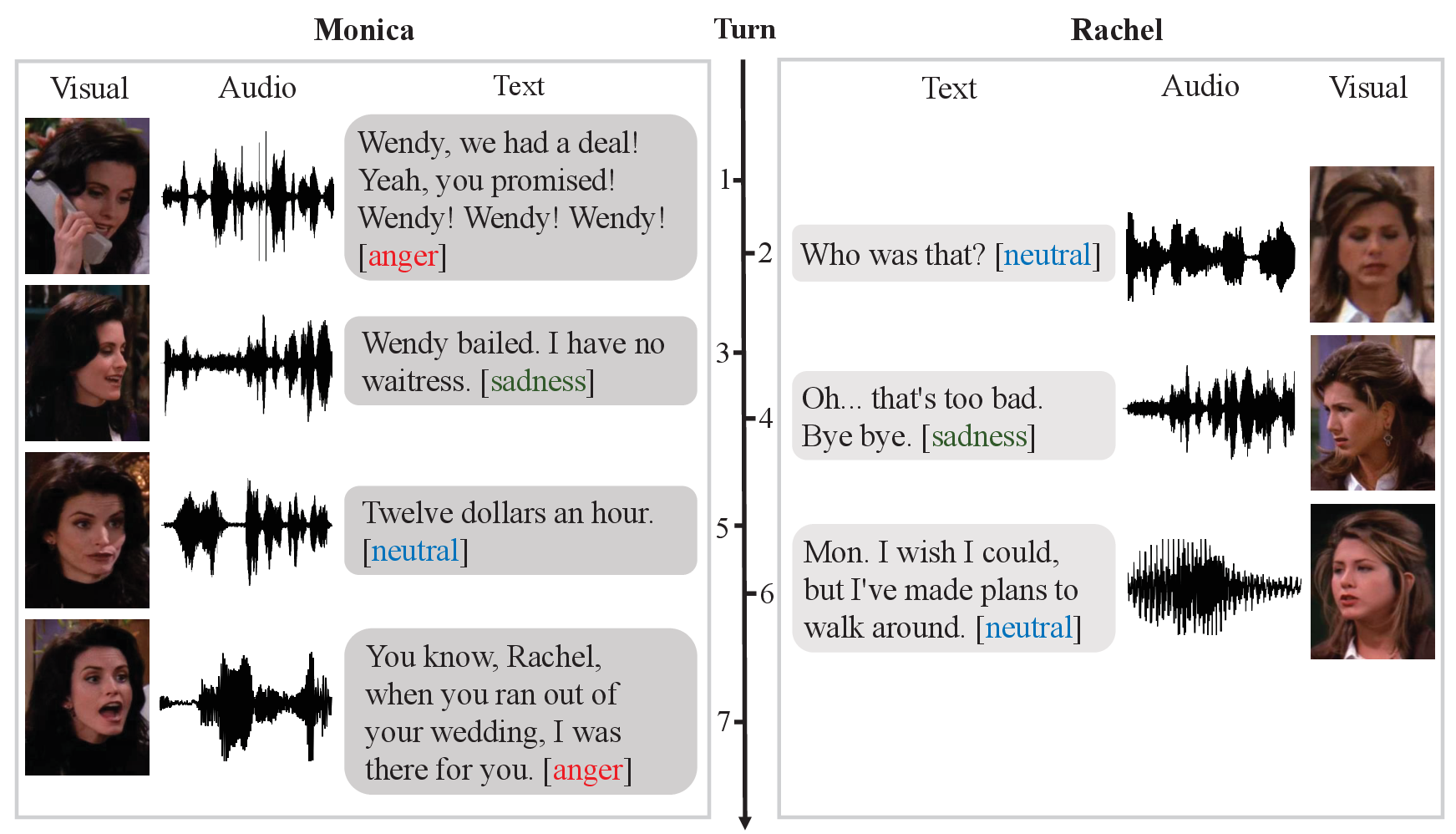}
\caption{A multimodal conversation example from the Friends TV series.}
\label{fig:1}
\end{figure}

Unlike emotion recognition in textual conversations, we argure that three key characteristics are essential for multimodal ERC: intra- and inter-modal interactions between conversation utterances, different contributions of modalities, and efficient modal representations. An example is shown in Fig.~\ref{fig:1}. (1) To understand the importance of intra- and inter-modal interactions, let us focus on the single and multiple modalities, respectively. Recognizing ``anger" emotion of the $7$th utterance spoken by Monica is difficult using only ``You know, Rachel, when you ran out of your wedding, I was there for you.", but it becomes easy when looking back to the textual expression of the $6$th utterance because Rachel has made plans to walk around. Additionally, we believe there are two types of inter-modal interactions: interactions between the same and different utterances. First, as stated above, it is hard to identify the emotion of the $7$th utterance using its textual expression; however, it also would be easy when fused with its visual and acoustic expressions since they burst instantaneously. Second, we know that the textual expression of the $5$th utterance shows ``neutral" emotion, and hence it could be possible to identify ``neutral" emotion of the $6$th utterance by interacting this utterance's visual expression and the $5$th utterance's textual expression. (2) To understand the importance of contributions of different modalities, let us focus on the $3$rd utterance. The textual and acoustic expressions play more critical roles in recognizing ``sadness" emotion than the visual expression because a smiling face usually means ``joy" emotion. (3) To understand the importance of efficient modal representations, let us focus on the textual expression of the $1$st utterance, which contains multiple exclamation points. If the learned representation does not include the meaning of ``!", it is challenging to identify ``anger" emotion.

Therefore, it is valuable to capture intra- and inter-modal interactions between utterances, dynamically learn weights between modalities, and enhance modal representations for multimodal ERC. However, existing studies of the task have some limitations in achieving these characteristics. On the one hand, most methods have drawbacks in modeling intra- and inter-modal interactions. For example, CMN \cite{hazarika:cmn2018}, ICON \cite{hazarika:icon2018}, and DialogueRNN \cite{majumder:dialoguernn2019} concatenate unimodal features at the input level, and thus cannot capture intra-modal interactions explicitly. While DialogueTRM \cite{mao:dialoguetrm2021} designs hierarchical transformer and multi-grained interaction fusion modules to explore intra- and inter-modal emotional behaviors, it ignores inter-modal interactions between different utterances. MMGCN \cite{hu:mmgcn2021} and MM-DFN \cite{hu2022mm} are graph-based fusion methods that require manually constructed graph structures to represent conversations. On the other hand, existing methods rely on the designed model to learn modal representations, but no work focuses on further improving modal representations using model-agnostic techniques for ERC.

In this work, a transformer-based model with self-distillation (SDT) is proposed to take into account the three aforementioned characteristics. First, we introduce intra- and inter-modal transformers in a modality encoder to capture intra- and inter-modal interactions, and take positional and speaker embeddings as additional inputs of these transformers to capture contextual and speaker information. Next, a hierarchical gated fusion strategy is proposed to dynamically fuse information from multiple modalities. Then, we predict emotion labels of conversation utterances based on fused multimodal representations in an emotion classifier. We call the above three components a transformer-based model. Finally, to learn more effective modal representations, we introduce self-distillation into the proposed transformer-based model, which transfers knowledge of hard and soft labels from the model to each modality. We treat the proposed model as the teacher and design three students according to three existing modalities. These students are trained by distilling knowledge from the teacher to learn better modal representations.

In summary, our contributions are as follows:
\renewcommand{\labelitemi}{$\bullet$}
\begin{itemize}
\item We propose a transformer-based model for multimodal ERC that contains a modality encoder for capturing intra- and inter-modal interactions between conversation utterances and a hierarchical gated fusion strategy for adaptively learning weights between modalities. 
\item To learn more effective modal representations, we devise self-distillation that transfers knowledge of hard and soft labels from the proposed model to each modality.
\item Experiments on two benchmark datasets show the superiority of our proposed model. In addition, several studies are conducted to investigate the impact of positional and speaker embeddings, intra- and inter-modal transformers, self-distillation loss functions, and hierarchical gated fusion strategy.
\end{itemize}

The rest of this paper is organized as follows: Section~\ref{sec:2} discusses the related work; Section~\ref{sec:3} formalizes the task definition and describes the proposed model; Section~\ref{sec:4} gives the experimental settings; Section~\ref{sec:5} presents the experimental results and discussion; Finally, Section~\ref{sec:6} concludes the paper and provides directions for further work.

\section{related work}
\label{sec:2}
\subsection{Emotion Recognition in Conversations}
ERC has attracted widespread interest among researchers with the increase in available conversation datasets, such as IMEOCAP \cite{busso:iemocap2008}, AVEC \cite{schuller:avec2012}, and MELD \cite{poria:meld2019}, etc. Early studies primarily used lexicon-based methods \cite{Lee:toward2005,devillers:real2006}. Recent works have generally resorted to deep neural networks and focused on modeling context- and speaker-sensitive dependencies. We divide the existing methods into two categories: speaker-ignorant and speaker-dependent methods, according to whether they utilize speaker information. 

Speaker-ignorant methods do not distinguish speakers and focus only on capturing contextual information in a conversation. HiGRU \cite{jiao:higru2019} contains two gated recurrent units (GRUs) to model contextual relationships between words and utterances, respectively. AGHMN \cite{jiao2020real} uses a hierarchical memory network to enhance utterance representations and introduce an attention GRU to model contextual information. MVN \cite{ma:mvn2022} utilizes a multi-view network to model word- and utterance-level dependencies in a conversation. In contrast, speaker-dependent methods model both context- and speaker-sensitive dependencies. DialogueRNN \cite{majumder:dialoguernn2019} leverages three distinct GRUs to update speaker, context, and emotional states in a conversation, respectively. DialogueGCN \cite{ghosal:dialoguegcn2019} uses a graph convolutional network to model speaker and conversation sequential information. HiTrans \cite{li:hitrans2020} consists of two hierarchical transformers to capture global contextual information and exploits an auxiliary task to model speaker-sensitive dependencies. 

However, most of them are proposed for the textual modality, ignoring the effectiveness of other modalities. Due to the promising performance in the multimodal community, some approaches tend to address multimodal ERC. DialogueTRM \cite{mao:dialoguetrm2021} explores intra- and inter-modal emotional behaviors using hierarchical transformer and multi-grained interaction fusion modules, respectively. MMGCN \cite{hu:mmgcn2021} constructs a fully connected graph to model multimodal and long-distance contextual information, and speaker embeddings are added for encoding speaker information. MM-DFN \cite{hu2022mm} designs a graph-based dynamic fusion module to reduce redundancy and enhance complementarity between modalities. MMTr \cite{zou2022improving} preserves the integrity of main modal representations and enhances weak modal representations by using multi-head attention. UniMSE \cite{hu2022unimse} performs modality fusion at syntactic and semantic levels and introduces inter-modality contrastive learning to differentiate fusion representations among samples. This paper focuses on exploring intra- and inter-modal interactions between utterances, learning weights between modalities, and enhancing modal representations for multimodal ERC.

\subsection{Multimodal Language Analysis}
Multimodal language analysis is a rapidly growing field and includes various tasks \cite{liang:multimodal2018}, such as multimodal emotion recognition, sentiment analysis, and personality traits recognition. The key in this area is to fuse multimodal information. Early studies on multimodal fusion mainly included early fusion and late fusion. Early fusion \cite{wollmer:youtube2013,poria:cmkl2016} integrates features of different modalities at the input level. Late fusion \cite{Nojavanasghari:deepmulti2016,kampman:investigating2018} constructs distinct models for each modality and then ensembles their outputs by majority voting or weighted averaging, etc. Unfortunately, as stated in  \cite{liu:efficient-low2018}, these two kinds of fusion methods cannot effectively capture intra- and inter-modal interactions.

Subsequently, model fusion has become popular and various models have been proposed. TFN \cite{zadeh:tensor2017} models unimodal, bimodal, and trimodal interactions explicitly by computing Cartesian product. LMF \cite{liu:efficient-low2018} utilizes low-rank weight tensors for multimodal fusion, which reduces the complexity of TFN. MFN \cite{ zadeh2018memory} learns cross-modal interactions with an attention mechanism and stores information over time by a multi-view gated memory. MulT \cite{tsai:multimodal2019} utilizes cross-modal transformers to model long-range dependencies across modalities. Rahman et al. \cite{rahman-etal-2020-integrating} fine-tuned large pre-trained transformer models for multimodal language by designing a multimodal adaptation gate (MAG). Self-MM \cite{yu2021learning} uses a unimodal label generation strategy to acquire independent unimodal supervision and then learns multimodal and unimodal tasks jointly. Yuan et al. \cite{yuan:trf2021} adopted transformer encoders to model intra- and inter-modal interactions between modality sequences. In order to capture intra- and inter-modal interactions between conversation utterances and meanwhile learn weights between modalities, we present a transformer-based model.

\subsection{Knowledge Distillation}
Knowledge distillation (KD) aims at transferring knowledge from a large teacher network to a small student network. The knowledge mainly includes soft labels of the last output layer (i.e., output-based knowledge) \cite{hinton:distilling2015}, features of intermediate layers (i.e., feature-based knowledge) \cite{romero2014fitnets}, and relationships between different layers (i.e., relation-based knowledge) \cite{Yim:gift2017}. Depending on the learning schemes, existing methods on KD are categorized into three classes: offline distillation \cite{passalis2018learning,li2020few}, online distillation
\cite{zhang2018deep,chung2020feature}, and self-distillation
\cite{zhang:self2019,hou2019learning}. In offline distillation, the teacher network is first trained and then the pre-trained teacher distills its knowledge to guide the student training. In online distillation, the teacher and student networks are updated simultaneously, and hence its training process is only one-phase.
Self-distillation is a special case of online distillation that teaches a single network using its own
knowledge.

Recently, KD has been used for multimodal emotion recognition. For example, Albanie et al. \cite{albanie2018emotion} transferred visual knowledge into a speech emotion recognition model using unlabelled video data. Wang et al. \cite{wang2020implicit} proposed K-injection subnetworks to distill linguistic and acoustic knowledge representing group emotions and transfer implicit knowledge into the audiovisual model for group emotion recognition. Schoneveld et al. \cite{schoneveld2021leveraging} applied KD to further improve performance for facial expression recognition. Most existing models belong to offline distillation, which requires training a teacher network. In contrast, self-distillation needs no extra network except for the network itself. While self-distillation has been successfully applied in computer vision and natural language processing \cite{moriya2020self,zhou2021automatic,luo:boost2021}, it focuses on unimodal tasks.

In this work, we adopt the idea of self-distillation to enhance modal representations for multimodal ERC. Moreover, output-based knowledge is used only due to the following reasons: (1) Soft labels can be used as training supervision which contain dark knowledge \cite{hinton:distilling2015} and can provide effective regularization for the model \cite{yuan2020revisiting}. (2) Intuitively, the features of different modalities vary widely, and hence matching fused multimodal features with unimodal features is inappropriate\footnote{We tried to add feature-based knowledge, but the performance drops significantly.}. (3) Our teacher and student networks lying in the same model have different architectures that results in an inability to inject relationships between different layers of the teacher network into the student network \cite{passalis2018learning}. Therefore, we adopt output-based knowledge rather than feature- and relation-based knowledge.

\section {Methodology}
\label{sec:3}
\subsection {Task Definition}
A conversation is composed of $N$ consecutive utterances $\left\{ {{u_1},{u_2},\cdots,{u_N}} \right\}$ and $M$ speakers $\left\{ {{s_1},{s_2}, \cdots ,{s_M}} \right\}$. Each utterance $u_i$ is spoken by a speaker ${s_{\phi \left( {u_i} \right)}}$, where $\phi $ is the mapping between an utterance and its corresponding speaker's index. Moreover, $u_i$ involves textual ($t$), acoustic ($a$), and visual ($v$) modalities, and their feature representations are denoted as ${\bf{u}}_i^t \in \mathbb{R}^{d_t}$, ${\bf{u}}_i^a \in \mathbb{R}^{d_a}$, and ${\bf{u}}_i^v \in \mathbb{R}^{d_v}$, respectively. We represent textual, acoustic, and visual modality sequences of all utterances in the conversation as ${{\bf{U}}_t} = \left[ {\bf{u}}_1^t;{\bf{u}}_2^t;\cdots; {\bf{u}}_N^t \right]\in \mathbb{R}^{N\times d_t}$, ${{\bf{U}}_a} = \left[ {\bf{u}}_1^a;{\bf{u}}_2^a;\cdots;{\bf{u}}_N^a \right]\in \mathbb{R}^{N\times d_a}$, and ${{\bf{U}}_v} = \left[ {\bf{u}}_1^v;{\bf{u}}_2^v;\cdots; {\bf{u}}_N^v \right]\in \mathbb{R}^{N\times d_v}$, respectively. The ERC task aims to predict the emotion label of each utterance $u_i$ from pre-defined emotion categories.

\subsection{Overview}
\begin{figure*}
	\centering
	\includegraphics[width=6.1in]{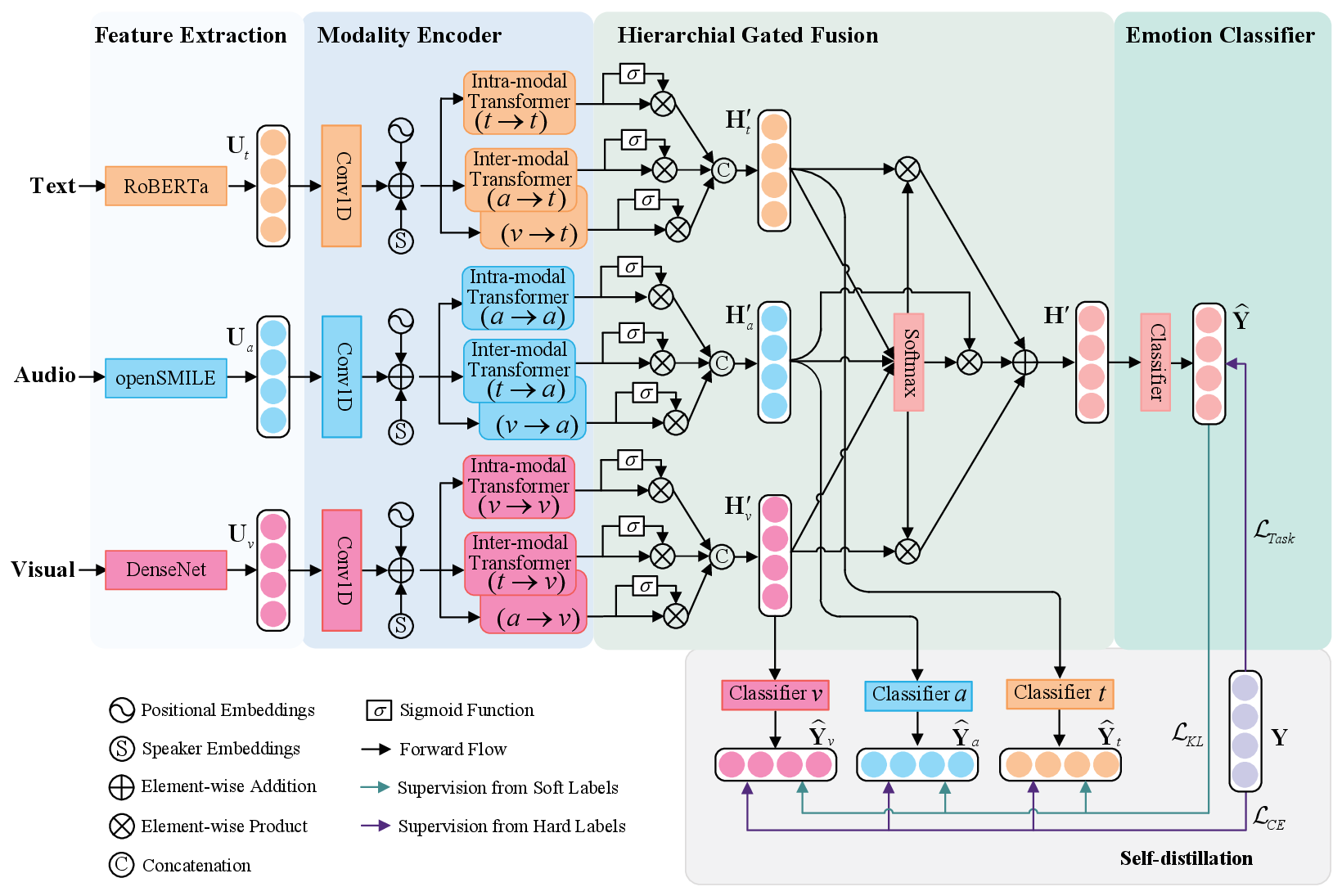}
	\caption{The overall architecture of SDT. After extracting utterance-level unimodal features, it consists of four key components: Modality Encoder, Hierarchical Gated Fusion, Emotion Classifier, and Self-distillation.}
	\label{fig:2}
\end{figure*}

Fig.~\ref{fig:2} gives an overview of our proposed SDT. After extracting utterance-level unimodal features, the transformer-based model consists of three modules: a modality encoder module for capturing intra- and inter-modal interactions between different utterances, a hierarchical gated fusion module for adaptively learning weights between modalities, and an emotion classifier module for predicting emotion labels. Furthermore, we introduce self-distillation and devise two kinds of losses to transfer knowledge from our proposed model within each modality to learn better modal representations.

\subsection{Modality Encoder}
The modality encoder obtains modality-enhanced modality sequence representations that can learn intra- and inter-modal interactions between conversation utterances.

\subsubsection*{\bf Temporal Convolution} To ensure that three unimodal sequence representations lie in the same space, we feed them into a 1D convolutional layer:
\begin{equation}
{\bf{U}}_{m}^{\prime}=\mathrm{Conv1D}\left({\bf{U}}_{m}, k_{m}\right) \in \mathbb{R}^{N\times d}, \, m \in\{t, a, v\},
\end{equation}
where $k_{m}$ is the size of convolutional kernel for $m$ modality, $N$ is the number of utterances in the conversation, and $d$ is the common dimension. 

\subsubsection*{\bf Positional Embeddings} To utilize positional and sequential information of the utterance sequence, we introduce positional embeddings \cite{vaswani:attention2017} to augment the convolved sequence:
\begin{equation}
\begin{array}{c}
{\bf{P E}}_{(p o s, 2 i)} =\sin \left(\frac{p o s}{10000^{2 i / d}}\right), \\\\
{\bf{P E}}_{(p o s, 2 i+1)} =\cos \left(\frac{p o s}{10000^{2 i / d}}\right),
\end{array}
\end{equation}
where $pos$ is the utterance index and $i$ is the dimension index.

\subsubsection*{\bf Speaker Embeddings} To capture speaker information of the utterance sequence, we also design speaker embeddings to augment the convolved sequence. Speaker ${s_j}$ in conversations is mapped into a vector:
\begin{equation}
	{{\bf{s}}_j} = {{\bf{V}}_s}{\bf{o}}\left( {{s_j}} \right) \in\mathbb{R}^{d},\;\;j = 1,\;2,\; \cdots ,\;M,
\end{equation}
where $M$ is the total number of speakers, ${{\bf{V}}_s} \in\mathbb{R}^{d\times M}$ is a trainable speaker embedding matrix, and ${\bf{o}}\left( {{s_j}} \right) \in\mathbb{R}^{M}$ is a one-hot vector of speaker ${s_j}$, i.e., $1$ in the $j$th position and $0$ otherwise. 

Hence, speaker embeddings corresponding to the conversation can be represented as ${\bf{SE}}= \left[ {\bf{s}}_{\phi \left( {u_1} \right)};{\bf{s}}_{\phi \left( {u_2} \right)};\cdots; {\bf{s}}_{\phi \left( {u_N} \right)} \right]$.

Overall, we augment positional and speaker embeddings to the convolved sequence:
\begin{equation}
    {\bf{H}}_{m}={\bf{U}}_{m}^{\prime}+{\bf{PE}}+{\bf{SE}}.
\end{equation}
\vspace{-2pt}
Here, ${\bf{H}}_{m}$ is the low-level positional- and speaker-aware utterance sequence representation for $m$ modality.

\subsubsection*{\bf Intra- and Inter-modal Transformers} We introduce intra- and inter-modal transformers to model intra- and inter-modal interactions for the utterance sequence, respectively. These transformers adopt the transformer encoder \cite{vaswani:attention2017}, which contains three inputs, queries ${\bf{Q}}\in\mathbb{R}^{T_q\times d_k}$, keys ${\bf{K}}\in\mathbb{R}^{T_k\times d_k}$, and values ${\bf{V}}\in\mathbb{R}^{T_k\times d_v}$. We denote the transformer encoder as $\mathrm{Transformer}\left({\bf{Q}}, {\bf{K}},{\bf{V}}\right)$.

For the intra-modal transformer, we take ${\bf{H}}_{m}$ as queries, keys, and values:
\begin{equation}
    {\bf{H}}_{m 
\to m}=\mathrm{Transformer}\left({\bf{H}}_{m}, {\bf{H}}_{m},{\bf{H}}_{m} \right) \in \mathbb{R}^{N\times d},
\end{equation}
where $m \in\{t, a, v\}$. The intra-modal transformer enhances $m$-modality sequence representation by itself and thus can capture intra-modal interactions between the utterance sequence.

For the inter-modal transformer, we take ${\bf{H}}_{m}$ as queries, and ${\bf{H}}_{n}$ as keys and values:
\begin{equation}
    {\bf{H}}_{n 
\to m}=\mathrm{Transformer}\left({\bf{H}}_{m}, {\bf{H}}_{n},{\bf{H}}_{n} \right) \in \mathbb{R}^{N\times d},
\end{equation}
where $m \in\{t, a, v\}$ and $n \in\{t, a, v\}-\{m\}$.
The inter-modal transformer enables $m$ modality to get information from $n$ modality and hence can capture inter-modal interactions between the utterance sequence.

In summary, $n$-enhanced $m$-modality sequence representation, ${\bf{H}}_{n \to m}$, is obtained from the modality encoder module, where $n, m \in\{t, a, v\}$.
\subsection{Hierarchical Gated Fusion}
We design a hierarchical gated fusion module containing unimodal- and multimodal-level gated fusions to adaptively obtain enhanced single-modality sequence representation and dynamically learn weights between these enhanced modality representations, respectively.

\subsubsection*{\bf Unimodal-level Gated Fusion} We first use a gated mechanism to filter out irrelevant information in ${\bf{H}}_{n
\to m}$:
\begin{equation}
    {g}_{n \to m} = \sigma \left ( {\bf{W}}_{n
\to m}\cdot {\bf{H}}_{n \to m}\right),
\end{equation}
\vspace{-15pt}
\begin{equation}
    {\bf{H}}_{n \to m}^{\prime}={\bf{H}}_{n 
\to m} \otimes {g}_{n \to m},
\end{equation}
where ${\bf{W}}_{n
\to m}\in \mathbb{R}^{d\times d}$ is a weight matrix, $\sigma$ is the sigmoid function, $\otimes$ is the element-wise product, and ${g}_{n \to m}$ denotes the gate.

Then, we concatenate ${\bf{H}}_{m 
\to m}^{\prime}$, ${\bf{H}}_{n_{1} 
\to m}^{\prime}$, and ${\bf{H}}_{n_{2}
\to m}^{\prime}$, followed by a fully connected (FC) layer to obtain enhanced $m$-modality sequence representation:
\begin{equation}
   {\bf{H}}_{m}^{\prime}\!=\!{\bf{W}}_{m}\!\cdot \!\left[{\bf{H}}_{m 
\to m}^{\prime}; {\bf{H}}_{n_{1} 
\to m}^{\prime}; {\bf{H}}_{n_{2}
\to m}^{\prime}\right] \!+ \! {\bf{b}}_{m}  \in \mathbb{R}^{N\times d},
\end{equation}
where $m \in\{t, a, v\}$, $n_1$ and $n_2$ represent other two modalities, ${\bf{W}}_{m}\in \mathbb{R}^{3d\times d}$ and ${\bf{b}}_{m}\in \mathbb{R}^{d}$ are trainable parameters.

We set ${\bf{H}}_{m}^{\prime}=\left[{\bf{h}}_{m1}^{\prime};{\bf{h}}_{m2}^{\prime};\cdots;{\bf{h}}_{mN}^{\prime}\right]$, where ${\bf{h}}_{mi}^{\prime}$ is enhanced $m$-modality representation for the utterance $u_i$.

\subsubsection*{\bf Multimodal-level Gated Fusion} We also design a gated mechanism using the softmax function to dynamically learn weights between enhanced modalities for each utterance.

Specifically, the final multimodal representation of the utterance $u_i$ is calculated by:
\begin{equation}
    \left[{ {\bf{g}}_{ti}; {\bf{g}}_{ai}; {\bf{g}}_{vi} }\right] = \mathrm{softmax}\left(\! {\left[ {{\bf{W}}\!\cdot\!{\bf{h}}_{ti}^{\prime};{\bf{W}}\!\cdot\!{\bf{h}}_{ai}^{\prime}; {\bf{W}}\!\cdot\! {\bf{h}}_{vi}^{\prime}} \right]} \!\right)\!,
\end{equation}
\vspace{-12pt}
\begin{equation}
    {\bf{h}}_{i}^{\prime}=\sum\limits_{m \in \left\{ {t,a,v} \right\}} {\bf{h}}_{mi}^{\prime} \otimes {\bf{g}}_{mi},
\end{equation}
where ${\bf{W}}\in \mathbb{R}^{d\times d}$ is a weight matrix, ${\bf{g}}_{ti}$, ${\bf{g}}_{ai}$, and ${\bf{g}}_{vi}$ are learned weights of $t$, $a$, $v$ modalities for the utterance $u_i$, respectively.

Thus, multimodal sequence representation of conversation utterances is obtained and denoted as ${\bf{H}}^{\prime}=\left[{\bf{h}}_{1}^{\prime};{\bf{h}}_{2}^{\prime};\cdots;{\bf{h}}_{N}^{\prime} \right]$.

\subsection{Emotion Classifier} To calculate probabilities over $C$ emotion categories, ${\bf{H}}^{\prime}$ is fed into a classifier with an FC and softmax layer:
\begin{equation}
    {\bf{E}}={\bf{W}}_{e} \cdot {\bf{H}}^{\prime}+
   {\bf{b}}_{e}\in \mathbb{R}^{N\times C},
\end{equation}
\vspace{-15pt}
\begin{equation}
   \hat{{\bf{Y}}} =\mathrm{softmax}\left(\bf{E}\right),
\end{equation}
where ${\bf{W}}_{e}\in \mathbb{R}^{d\times C}$ and ${\bf{b}}_{e}\in \mathbb{R}^{C}$ are trainable parameters. We set $\hat{{\bf{Y}}}=\left[\hat{{\bf{y}}}_{1};\hat{{\bf{y}}}_{2};\cdots;\hat{{\bf{y}}}_{N}\right]$, where $\hat{{\bf{y}}}_{i}$ is the emotion probability vector for the utterance $u_i$. Finally, we choose $\operatorname{argmax}\left(\hat{{\bf{y}}}_{i}\right)$ as the predicted emotion label for $u_i$.

\subsubsection*{\bf Task Loss} We utilize the cross-entropy loss for estimating the quality of emotion predictions during training:
\begin{equation}
    \mathcal{L}_{Task} =  - \frac{1}{N}\sum\limits_{i = 1}^N {\sum\limits_{j = 1}^C {{{\bf{y}}_{i,j}}\log \left( {{{\widehat {\bf{y}}}_{i,j}}} \right)} } ,
\end{equation}
where $N$ represents the number of utterances in the conversation, and $C$ represents the number of emotion classes. ${{\bf{y}}_i}$ and ${{{\widehat {\bf{y}}}_i}}$ denote the ground-truth one-hot vector and probability vector for the emotion of $u_i$, respectively.

\subsection{Self-distillation}
Soft labels containing informative dark knowledge can be used as training supervision; hence, we devise self-distillation to transfer knowledge of hard and soft labels to each modality, and guide the model in learning more expressive modal representations.

We treat our proposed transformer-based model as the teacher and design three students according to existing modalities. Specifically, a classifier consisting of an FC and softmax layer only used during training, is set after each unimodal-level gated fusion. During training, textual, acoustic, and visual modality encoders with their corresponding unimodal-level gated fusions and classifiers are trained as three students (i.e., student $t$, student $a$, student $v$) via distilling from the teacher. 

The output of student $m$ is its predicted emotion probabilities:
\begin{equation}
    {\bf{E}}_{m}={\bf{W}}_{m}^{\prime} \cdot \mathrm{ReLU}\left({\bf{H}}_{m}^{\prime} \right)+
   {\bf{b}}_{m}^{\prime}\in \mathbb{R}^{N\times C},
\end{equation}
\begin{equation}
\begin{array}{c}
   \hat{{\bf{Y}}}_{m} =\mathrm{softmax}\left({\bf{E}}_{m}\right),\\
  \hat{{\bf{Y}}}_{m}^{\tau} =\mathrm{softmax}\left({\bf{E}}_{m}/\tau\right),
\end{array}
\end{equation}
where $m \in\{t, a, v\}$, ${\bf{W}}_{m}^{\prime}\in \mathbb{R}^{d\times C}$ and ${\bf{b}}_{m}^{\prime}\in \mathbb{R}^{C}$ are trainable parameters. $\tau$ is the temperature to soften $\hat{{\bf{Y}}}_{m}$ (written as $\hat{{\bf{Y}}}_{m}^{\tau}$ after softened) and a higher $\tau$ produces a softer distribution over classes \cite{hinton:distilling2015}. We set $\hat{{\bf{Y}}}_{m}=\left[\hat{{\bf{y}}}_{m1};\hat{{\bf{y}}}_{m2};\cdots;\hat{{\bf{y}}}_{mN}\right]$ and $\hat{{\bf{Y}}}_{m}^{\tau}=\left[\hat{{\bf{y}}}_{m1}^{\tau};\hat{{\bf{y}}}_{m2}^{\tau};\cdots;\hat{{\bf{y}}}_{mN}^{\tau}\right]$.

During training, we introduce two kinds of losses to train the student $m$ to learn better enhanced $m$-modality sequence representation, where $m \in\{t, a, v\}$.

\subsubsection*{\bf Cross Entropy Loss} We minimize the cross entropy loss between the predicted probability of the student $m$ and the ground-truth:
\begin{equation}
    \mathcal{L}_{CE}^{m}= - \frac{1}{N}\sum\limits_{i = 1}^N {\sum\limits_{j = 1}^C {{{\bf{y}}_{i,j}}\log \left( {{{\widehat {\bf{y}}}_{mi,j}}} \right)} },
\end{equation}
where ${\widehat {{\bf{y}}}_{mi}}$ is the emotion probability vector of the student $m$ for $u_i$. In this way, knowledge from hard labels is directly introduced to the student to learn better modal representations.

\subsubsection*{\bf KL Divergence Loss} To make the output probability of the student $m$ approximate the output of the teacher (i.e., soft labels), the Kullback-Leibler (KL) divergence loss between them is minimized:
\begin{equation}
    \mathcal{L}_{KL}^{m} = \frac{1}{N}\sum\limits_{i = 1}^N {\sum\limits_{j = 1}^C {{{\widehat {\bf{y}}}_{mi,j}^{\tau}}\log \left( {\frac{{{{\widehat {\bf{y}}}_{mi,j}^{\tau}}}}{{{{\widehat {\bf{y}}}_{i,j}^{\tau}}}}} \right)} },
\end{equation}
where ${{\widehat {\bf{y}}}_{mi}}^{\tau}$ and ${{\widehat {\bf{y}}}_{i}}^{\tau}$ are soften probability distributions of the student $m$ and the teacher, respectively. In this way, knowledge from soft labels is transferred to the student to learn better modal representations.

With both hard and soft labels, the overall loss can be expressed as:
\begin{equation}
    \mathcal{L} = \gamma_1 \mathcal{L}_{Task} + \gamma_2 \mathcal{L}_{CE} + \gamma_3 \mathcal{L}_{KL},
\end{equation}
\begin{equation}
\mathcal{L}_{CE} = \sum\limits_{m \in \left\{ {t,a,v} \right\}} \mathcal{L}_{CE}^m,
\end{equation}
\begin{equation}
\mathcal{L}_{KL} = \sum\limits_{m \in \left\{ {t,a,v} \right\}} \mathcal{L}_{KL}^m,
\end{equation}
where $\gamma_1$, $\gamma_2$, and $\gamma_3$ are hyper-parameters that control the weights of the three kinds of losses. In experiments, we set $\gamma_1=\gamma_2=\gamma_3=1$.

\section{Experimental Settings}
\label{sec:4}
\subsection{Datasets and Evaluations}
We use IEMOCAP \cite{busso:iemocap2008} and MELD \cite{poria:meld2019} datasets to evaluate the proposed model. The statistics of the two datasets are listed in Table~\ref{tab:1}.

\subsubsection*{\bf IEMOCAP} The dataset consists of two-way conversations of ten speakers, containing $153$ conversations and $7,433$ utterances. The dataset is divided into five sessions, where the first four sessions are used for training, while the last one is for testing. Each utterance is labeled with one of six emotions: happy, sad, neutral, angry, excited, and frustrated.

\subsubsection*{\bf MELD} This is a multi-speaker conversation dataset collected from the Friends TV series, containing $1,433$ conversations and $13,708$ utterances. Each utterance is labeled with one of seven emotions: neutral, surprise, fear, sadness, joy, disgust, and anger.

\begin{table}
\scriptsize
\centering
\caption{Statistics of the two datasets.}
\begin{tabular}{cccccccc}
\toprule
\multirow{2}{*}{Dataset} & \multicolumn{2}{c}{\#Conversations} & \multicolumn{2}{c}{\#Utterances} & \multirow{2}{*}{\#Classes} \\
\cmidrule(lr){2-3}\cmidrule(lr){4-5} 
& \multicolumn{1}{c}{Train+Val}        & \multicolumn{1}{c}{Test} & \multicolumn{1}{c}{Train+Val}      & \multicolumn{1}{c}{Test} & \\ 
\midrule
IEMOCAP & 120 & 31   & 5810 & 1623 & 6 \\
MELD    & 1153  & 280  & 11098 &  2610 & 7 \\ 
\bottomrule
\end{tabular}
\label{tab:1}
\end{table}

\subsubsection*{\bf Evaluation Metrics} Following previous works \cite{majumder:dialoguernn2019,ghosal:dialoguegcn2019}, we report the overall accuracy and weighted average F1-score to measure overall performance, and also present the accuracy and F1-score on each emotion class.

\subsection{Feature Extraction}
We extract utterance-level unimodal features as follows.
\subsubsection*{\bf Textual Modality} Following \cite{ghosal-etal-2020-cosmic}, we employ RoBERTa Large model \cite{liu2019roberta} to extract textual features. Roberta, a pre-trained model using a multi-layer transformer encoder architecture, builds on BERT which can efficiently learn textual representations. We fine-tune RoBERTa for emotion recognition from conversation transcripts and then take [\textsl{CLS}] tokens' embeddings at the last layer as textual features. The dimensionality of textual feature representation is $1024$.

\subsubsection*{\bf Acoustic Modality} Following \cite{hu:mmgcn2021}, we use openSMILE \cite{eyben2013recent} for acoustic feature extraction. openSMILE, a flexible feature extraction toolkit for signal processing, provides a scriptable console application to configure modular feature extraction components. After using openSMILE toolkit, an FC layer reduces the dimensionality of acoustic feature representation to $1582$ for IEMOCAP and $300$ for MELD.

\subsubsection*{\bf Visual Modality} Following \cite{hu:mmgcn2021}, we use DenseNet \cite{huang2017densely} pre-trained on Facial Expression Recognition Plus dataset for visual feature extraction. DenseNet, an effective CNN architecture, consists of multiple dense blocks, each of which contains multiple layers. The output of DenseNet is set to $342$; that is, the dimensionality of visual feature representation is $342$.

\subsection{Baselines}
We compare SDT with the following baseline models.
\subsubsection*{\bf CMN \cite{hazarika:cmn2018}} It uses two GRUs and memory networks to model contextual information for both speakers, but it is only available for dyadic conversations.

\subsubsection*{\bf ICON \cite{hazarika:icon2018}} It is an extension of CMN that captures inter-speaker emotional influences using another GRU. Similar to CMN, the model is applied to dyadic conversations.

\subsubsection*{\bf DialogueRNN \cite{majumder:dialoguernn2019}} It adopts three distinct GRUs to track the speaker, context, and emotional states in conversations, respectively.

The above models concatenate textual, acoustic, and visual features to obtain multimodal utterance representations. 

\subsubsection*{\bf MMGCN \cite{hu:mmgcn2021}} It constructs a conversation graph based on all three modalities and designs a multimodal fused graph convolutional network to model contextual dependencies across multiple modalities.

\subsubsection*{\bf DialogueTRM \cite{mao:dialoguetrm2021}} It uses a hierarchical transformer to manage the differentiated context preference within each modality and designs a multi-grained interactive fusion for learning different contributions across modalities for an utterance.

\subsubsection*{\bf MM-DFN \cite{hu2022mm}} It designs a graph-based dynamic fusion module to fuse multimodal context features, and this module could reduce redundancy and enhance complementarity between modalities.

\subsubsection*{\bf MMTr \cite{zou2022improving}} It uses distinct bidirectional long short-term memory networks (Bi-LSTMs) to learn contextual representations at the speaker’s self-context level and contextual context level, and designs a cross-modal fusion module to enhance weak modal representations.

\subsubsection*{\bf UniMSE \cite{hu2022unimse}} It uses T5 to fuse acoustic and visual modal features with multi-level textual features, and performs inter-modality contrastive learning to obtain discriminative multimodal representations.

For a fair comparison, we re-run all baselines, except MMTr and UniMSE, whose source codes are not released\footnote{We carefully implemented DialogueTRM to explore its performance using our extracted features, since its source code is not available; MMTr uses basically same feature extractors as us, and therefore we did not implement it; UniMSE uses T5 to learn contextual information on textual sequences and embeds multimodal fusion layers into T5, and hence our extracted features cannot be used for UniMSE and we also did not implement it.}. In addition, we re-implement DialogueRNN, MMGCN, DiaglogueTRM, and MM-DFN with our extracted features, namely DialogueRNN*, MMGCN*, DiaglogueTRM*, and MM-DFN*. We use the same data splits to implement all models.

\subsection{Implementation Details}
We implement the proposed model using Pytorch\footnote{\url{https://pytorch.org/}} and use Adam \cite{Kingma:adam2014} as optimizer with an initial learning rate of $1.0e-4$ for IEMOCAP and $5.0e-6$ for MELD. The batch size is $16$ for IEMOCAP and $8$ for MELD, and the temperature $\tau$ for the two datasets are set to $1$ and $8$, respectively. For the 1D convolutional layers, the number of input channels are set to $1024$, $1582$, and $342$ for textual, acoustic, and visual modalities, respectively (i.e., their corresponding feature dimensions) on IEMOCAP. On MELD, these parameters are set to $1024$, $300$, and $342$, respectively. In addition, the number of output channels and kernel size are set to $1024$ and $1$ respectively for all three modalities on the two datasets. For the transformer encoder, the hidden size, number of attention heads, feed-forward size, and number of layers are set to $1024$, $8$, $1024$, and $1$, respectively. To prevent overfitting, we set the L2 weight decay to $1.0e-5$ and employ dropout with a rate of $0.5$. All results are averages of $10$ runs.

\section{Results and Discussion}
\label{sec:5}
\subsection{Overall Results}
\begin{table*}[!t]
	\centering
	\caption{Results on the IEMOCAP dataset; ``*": baselines are re-implemented using our extracted features; bold font denotes the best performance.
	}
	\begin{tabular}{l|cccccccccccccc}
		\toprule
		\multirow{3}{*}{Models} & \multicolumn{14}{c}{IEMOCAP}                  \\ \cline{2-15} 
		& \multicolumn{2}{c}{happy} & \multicolumn{2}{c}{sad}  & \multicolumn{2}{c}{neutral} & \multicolumn{2}{c}{angry} & \multicolumn{2}{c}{excited} & \multicolumn{2}{c|}{frustrated} & \multirow{3}{*}{ACC} &
		\multirow{3}{*}{w-F1} \\ 
		\cmidrule(lr){2-3}\cmidrule(lr){4-5}\cmidrule(lr){6-7}\cmidrule(lr){8-9}\cmidrule(lr){10-11}\cmidrule(lr){12-13} &
		\multicolumn{1}{c}{ACC} &
		\multicolumn{1}{c}{F1} &
		\multicolumn{1}{c}{ACC} &
		\multicolumn{1}{c}{F1} &
		\multicolumn{1}{c}{ACC} &
		\multicolumn{1}{c}{F1} &
		\multicolumn{1}{c}{ACC} &
		\multicolumn{1}{c}{F1} &
		\multicolumn{1}{c}{ACC} &
		\multicolumn{1}{c}{F1} &
		\multicolumn{1}{c}{ACC} &
		\multicolumn{1}{c|}{F1} \\
		\midrule
		CMN & 24.31 & 30.30 & 56.33	& 62.02 & 52.34 & 52.41 &	61.76 & 60.17 & 56.19 & 60.76 & \textbf{72.44} & \multicolumn{1}{c|}{61.27} & 56.87 & 56.33 \\
		ICON & 25.00 & 31.30 & 67.35 &	73.17 & 55.99 & 58.50 & 69.41 &	66.29 & 70.90 &	67.09 & 71.92 & \multicolumn{1}{c|}{65.08} & 62.85 & 62.25 \\
		DialogueRNN & 25.00 & 34.95 & 82.86	& 84.58 & 54.43 &	57.66 &	61.76
		& 64.42 & \textbf{90.97} &	76.30 & 62.20 & \multicolumn{1}{c|}{59.55} & 65.43 & 64.29 \\
		MMGCN & 32.64 & 39.66 & 72.65 & 76.89	& 65.10 & 62.81 & 73.53	& \textbf{71.43}	& 77.93 & 75.40 & 65.09 & \multicolumn{1}{c|}{63.43} &  66.61 & 66.25 \\
		DialogueTRM & 61.11 & 57.89 & 84.90 & 81.25 & 69.27 & 68.56 & 76.47 & 65.99 & 76.25 & 76.13 & 50.39 & \multicolumn{1}{c|}{58.09} & 68.52 & 68.20 \\ 
		MM-DFN & 44.44 & 44.44 & 77.55 & 80.00 & 71.35 & 66.99 & 75.88 & 70.88 & 74.25 & 76.42 & 58.27 & \multicolumn{1}{c|}{61.67} & 67.84 & 67.85\\ 
		MMTr & - & - & - & - & - & - & - & - & - & - & - & \multicolumn{1}{c|}{-} & 72.27 & 71.91 \\ 
		UniMSE & - & - & - & - & - & - & - & - & - & - & - & \multicolumn{1}{c|}{-} & 70.56 & 70.66 \\ \hline
		DialogueRNN* & 57.64 & 57.64 &	77.96 & 80.25 &	75.52 & 70.56 & 68.24 &	64.99 & 73.91 &	75.95 & 59.06 & \multicolumn{1}{c|}{62.41} & 69.38 & 69.37 \\
		MMGCN* & 50.00 & 56.25 & 78.78 &	81.43 & 71.35 & 67.57 &	68.24 & 66.29 & 75.92
		&	76.82 &	65.09 & \multicolumn{1}{c|}{64.92} & 69.62 & 69.61 \\
		DialogueTRM* & 72.22 & 62.84 & \textbf{85.71} & 83.33 & 69.27 & 68.12 & \textbf{79.41} & 66.67 & 67.22 & 75.00 & 57.22 & \multicolumn{1}{c|}{63.28} & 69.87 & 69,93 \\
		MM-DFN* & 57.64 & 52.87 & 84.49 & \textbf{86.07} & 76.04 & 71.66 & 70.59 & 65.04 & 73.24 & 75.26 & 55.91 & \multicolumn{1}{c|}{62.19} & 69.87 & 69.91 \\ \hline
		SDT (Ours) & \textbf{72.71} & \textbf{66.19} & 79.51 &	81.84 & \textbf{76.33} & \textbf{74.62} & 71.88 &	69.73 & 76.79 & \textbf{80.17} & 67.14 & \multicolumn{1}{c|}{\textbf{68.68
		}} & \textbf{73.95} & \textbf{74.08} \\ 
		w/o self-distillation & 71.53 & 58.52 & 79.59 & 79.43 & 69.27 &	70.65 &	69.41
		& 67.05 & 69.23 &	77.09 & 67.98 & \multicolumn{1}{c|}{68.07} & 70.73 & 71.10 \\
		\bottomrule
	\end{tabular}
	\label{tab:2}
\end{table*}

\begin{table*}[!t]
	\centering 
	\caption{Results on the MELD dataset.}
	\resizebox{\textwidth}{!}{
	\begin{tabular}{l|cccccccccccccccc}
		\toprule3
		\multirow{3}{*}{Models} & \multicolumn{16}{c}{MELD}                  \\ \cline{2-17} 
		& \multicolumn{2}{c}{neutral} & \multicolumn{2}{c}{surprise} & \multicolumn{2}{c}{fear} & \multicolumn{2}{c}{sadness} & \multicolumn{2}{c}{joy} & \multicolumn{2}{c}{disgust} & \multicolumn{2}{c|}{anger} & \multirow{3}{*}{ACC} &
		\multirow{3}{*}{w-F1} \\
		\cmidrule(lr){2-3}\cmidrule(lr){4-5}\cmidrule(lr){6-7}\cmidrule(lr){8-9}\cmidrule(lr){10-11}\cmidrule(lr){12-13}\cmidrule(lr){14-15} &
		\multicolumn{1}{c}{ACC} &
		\multicolumn{1}{c}{F1} &
		\multicolumn{1}{c}{ACC} &
		\multicolumn{1}{c}{F1} &
		\multicolumn{1}{c}{ACC} &
		\multicolumn{1}{c}{F1} &
		\multicolumn{1}{c}{ACC} &
		\multicolumn{1}{c}{F1} &
		\multicolumn{1}{c}{ACC} &
		\multicolumn{1}{c}{F1} &
		\multicolumn{1}{c}{ACC} &
		\multicolumn{1}{c}{F1} &
		\multicolumn{1}{c}{ACC} &
		\multicolumn{1}{c|}{F1} \\
		\midrule
		DialogueRNN  & 82.17 & 76.56 & 46.62 & 47.64 & 0.00 & 0.00 & 21.15 & 24.65 & 49.50 & 51.49 & 0.00 & 0.00 & 48.41 & \multicolumn{1}{c|}{46.01} & 60.27 & 57.95 \\
		MMGCN & 84.32 & 76.96 & 47.33 & 49.63 &2.00 & 3.64 & 14.90 & 20.39 & 56.97 & 53.76 & 1.47 & 2.82 & 42.61 & \multicolumn{1}{c|}{45.23} & 61.34 & 58.41 \\
		DialogueTRM  & 83.20 & 79.41 & 56.94 & 55.27 & 12.00 & 17.39 & 27.88 & 36.48 & 60.45 & 60.30 & 16.18 & 20.18 & 51.01 & \multicolumn{1}{c|}{49.79} & 65.10 & 63.80 \\
		MM-DFN  & 79.06 & 75.80 & 53.02 & 50.42 & 0.00 & 0.00 & 17.79 & 23.72 & 59.20 & 55.48 & 0.00 & 0.00 & 50.43 & \multicolumn{1}{c|}{48.27} & 60.96 & 58.72 \\ 
		MMTr & - & - & - & - & - & - & - & - & - & - & - & - & - & \multicolumn{1}{c|}{-} & 64.64 & 64.41 \\
		UniMSE & - & - & - & - & - & - & - & - & - & - & - & - & - & \multicolumn{1}{c|}{-} & 65.09 & 65.51 \\ \hline
		DialogueRNN* & \textbf{85.11} & 79.60 & 54.09 & 56.72 & 10.00 & 12.66 & 29.81 & 38.63 & 62.94 & 63.81 & 22.06 & 27.27 & 53.62 & \multicolumn{1}{c|}{53.24} & 66.70 & 65.31 \\
		MMGCN*  & 81.53 & 79.20 & 58.36 & 57.75 & 8.00 & 13.79 & 31.73 & 39.40 & \textbf{69.90} & 63.43 & 20.59 & 24.56 & 52.17 & \multicolumn{1}{c|}{53.49} & 66.40 & 65.21 \\
		DialogueTRM*  & 83.44 & 79.54 & 54.45 & 57.09 & 24.00 & \textbf{27.91} & 33.17 & 40.95 & 60.45 & 62.79 & 22.06 & 28.04 & \textbf{58.26} & \multicolumn{1}{c|}{53.96} & 66.70 & 65.76 \\ 
		MM-DFN*  & 83.52 & 79.65 & \textbf{63.35} & 58.17 & \textbf{32.00} & 26.67 & 26.44 & 35.71 & 63.68 & \textbf{64.89} & 19.12 & 24.76 & 49.28 & \multicolumn{1}{c|}{52.15} & 66.55 & 65.48 \\ \hline
		SDT (Ours) & 83.22 & \textbf{80.19} & 61.28 & \textbf{59.07} & 13.80 & 17.88 & \textbf{34.90} & \textbf{43.69} & 63.24 & 64.29 & 22.65 & \textbf{28.78} & 56.93  & \multicolumn{1}{c|}{\textbf{54.33}} & \textbf{67.55} & \textbf{66.60} \\ 
		w/o self-distillation & 82.01 & 80.00 & 57.65 & 57.96 & 20.00 & 23.81 & 32.21  & 41.61 & 65.17 & 64.22 & \textbf{25.00} & 27.42 & 57.97 & \multicolumn{1}{c|}{54.05} & 66.97 & 66.26 \\
		\bottomrule
	\end{tabular}}
	\label{tab:3}
\end{table*}

Table~\ref{tab:2} and Table~\ref{tab:3} present the performance of baselines and SDT on IEMOCAP and MELD datasets, respectively. On IEMOCAP dataset, SDT performs better than all baselines and outperforms MMTr by $1.68\%$ and $2.17\%$ in terms of overall accuracy and weighted F1-score, respectively. In addition, SDT achieves a significant improvement on most emotion classes in terms of F1-score. On MELD dataset, SDT achieves the best performance compared to all baselines in terms of overall accuracy and weighted F1-score, and outperforms UniMSE by $2.46\%$ and $1.09\%$, respectively. Similar to IEMOCAP, SDT performs superior on most emotion classes in terms of F1-score.

Overall, the above results indicate the effectiveness of SDT. Furthermore, we have several similar findings on the two datasets: (1) DialogueTRM has a superior performance compared to DialogueRNN, MMGCN, and MM-DFN that use TextCNN \cite{kim-2014-convolutional} to extract textual features. This is because textual modality plays a more important role for ERC \cite{hu:mmgcn2021}, and DialogueTRM extracts textual features using BERT \cite{devlin-etal-2019-bert}, which is more powerful than TextCNN. (2) The baselines gain further improvement and achieve comparable results when using our extracted utterance features. The results show that our feature extractor is more effective and sequence- and graph-based baselines can achieve similar performance using our extracted features. (3) Even without self-distillation, our proposed model is still comparable to strong baselines, demonstrating the power of the proposed transformer-based model.

\subsection{Ablation Study}
We carry out ablation experiments on IEMOCAP and MELD. Table~\ref{tab:4} reports the results under different ablation settings.

\begin{table}[!t]
	\centering
	\caption{Results of ablation studies on the two datasets.}
	\begin{tabular}{lcccc}
		\toprule
		 & \multicolumn{2}{c}{IEMOCAP} & \multicolumn{2}{c}{MELD} \\
		 \cmidrule(lr){2-3}\cmidrule(lr){4-5}& 
		 \multicolumn{1}{c}{ACC} &
		 \multicolumn{1}{c}{w-F1} &
		 \multicolumn{1}{c}{ACC} &
		 \multicolumn{1}{c}{w-F1} \\
		 \midrule
		SDT & \textbf{73.95} & \textbf{74.08} & \textbf{67.55} & \textbf{66.60} \\ \hline
		Transformer-based model & & & & \\
		w/o positional embeddings & 72.27 & 72.39 & 66.86
		& 66.20 \\
		w/o speaker embeddings & 71.84 & 72.03 & 67.13 & 66.18 \\
		w/o intra-modal transformers & 73.38 & 73.36 & 67.13 & 66.21 \\
		w/o inter-modal transformers & 72.09 & 72.26 & 66.97 & 65.55 \\ \hline
		Self-distillation & & & & \\
		w/o $\mathcal{L}_{CE}$ & 73.07 & 73.32 & 67.39 & 66.37 \\
		w/o $\mathcal{L}_{KL}$ & 72.95 & 73.03 & 67.09 &  66.33 \\ \hline
		Modality & & & & \\
		Text & 66.42 & 66.58 & 66.82 & 65.52 \\
		Audio & 59.77 & 59.34 & 48.12 & 40.81 \\
		Visual & 41.47 & 42.71 & 48.05 & 32.01 \\
		Text + Audio & 72.52 & 72.75 & 67.05 & 66.24 \\
		Text + Visual & 69.01 & 69.07 & 67.20 & 66.18 \\
		Audio + Visual & 62.05 & 62.26 & 47.24 & 40.21 \\
		\bottomrule
	\end{tabular}
	\label{tab:4}
\end{table}

\subsubsection*{\bf Ablation on Transformer-based Model} Positional embeddings, speaker embeddings, intra-modal transformers, and inter-modal transformers are four crucial components of our proposed transformer-based model. We remove only one component at a time to evaluate the effectiveness of the component. From Table~\ref{tab:4}, we conclude that: (1) All components are useful because removing one of them leads to performance degradation. (2) Positional and speaker embeddings have considerable effects on the two datasets, which means capturing sequential and speaker information are valuable. (3) Inter-modal transformers are more important than intra-modal transformers on the two datasets. This indicates that inter-modal interactions between conversation utterances could provide more helpful information.

\subsubsection*{\bf Ablation on Self-distillation Loss Functions} There are two kinds of losses ( i.e., $\mathcal{L}_{CE}$ and $\mathcal{L}_{KL}$) for self-distillation. To verify the importance of these losses, we remove one loss at a time. Table~\ref{tab:4} shows that $\mathcal{L}_{CE}$ and $\mathcal{L}_{KL}$ are complementary and our model performs best when all losses are included. The result demonstrates that transferring knowledge of both hard and soft labels from the proposed transformer-based model to each modality can further boost the model performance.

\subsubsection*{\bf Effect of Different Modalities} To show the effect of different modalities, we remove one or two modalities at a time. From Table~\ref{tab:4}, we observe that: (1) For unimodal results, the textual modality has far better performance than the other two modalities, indicating that the textual feature plays a leading role in ERC. This finding is consistent with previous works \cite{mao:dialoguetrm2021,hu:mmgcn2021,hu2022mm}. (2) Any bimodal results are better than its own unimodal results. Moreover, fusing the textual modality and acoustic or visual modality performs superior to the fusion of the acoustic and visual modalities due to the importance of textual features. (3) Using all three modalities gives the best performance. The result can validate that emotion is affected by verbal, vocal, and visual expressions, and integrating multimodal information is essential for ERC.

\subsubsection*{\bf Effect of Different Fusion Strategies}

\begin{figure}[!t]
\hspace{-5mm}
\subfloat[Overall accuracy]{\label{fig:3a}
  \includegraphics[width=1.860in]{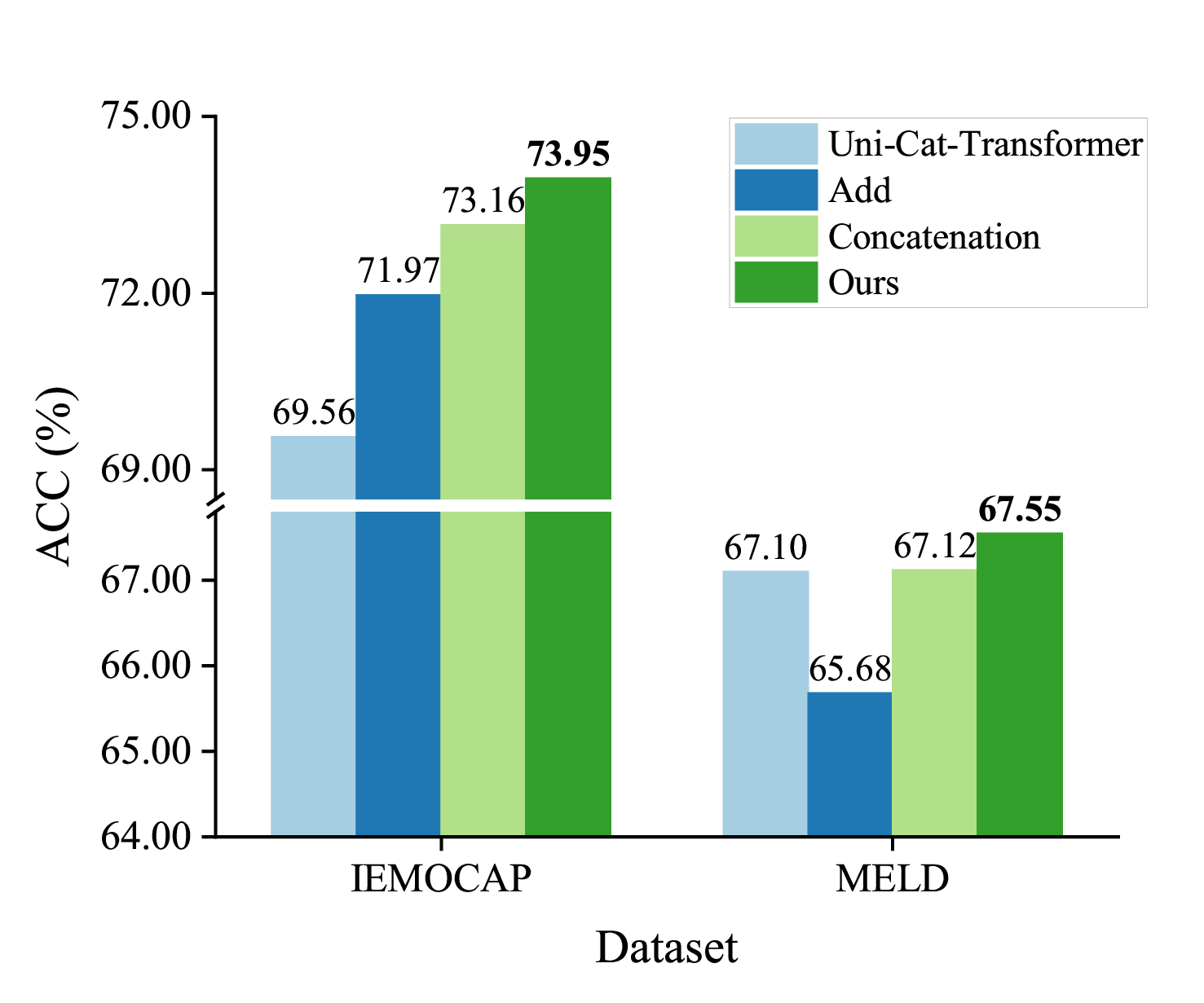}}
\hspace{-5.0mm}
\subfloat[Weighted F1-score]{\label{fig:3b}
  \includegraphics[width=1.860in]{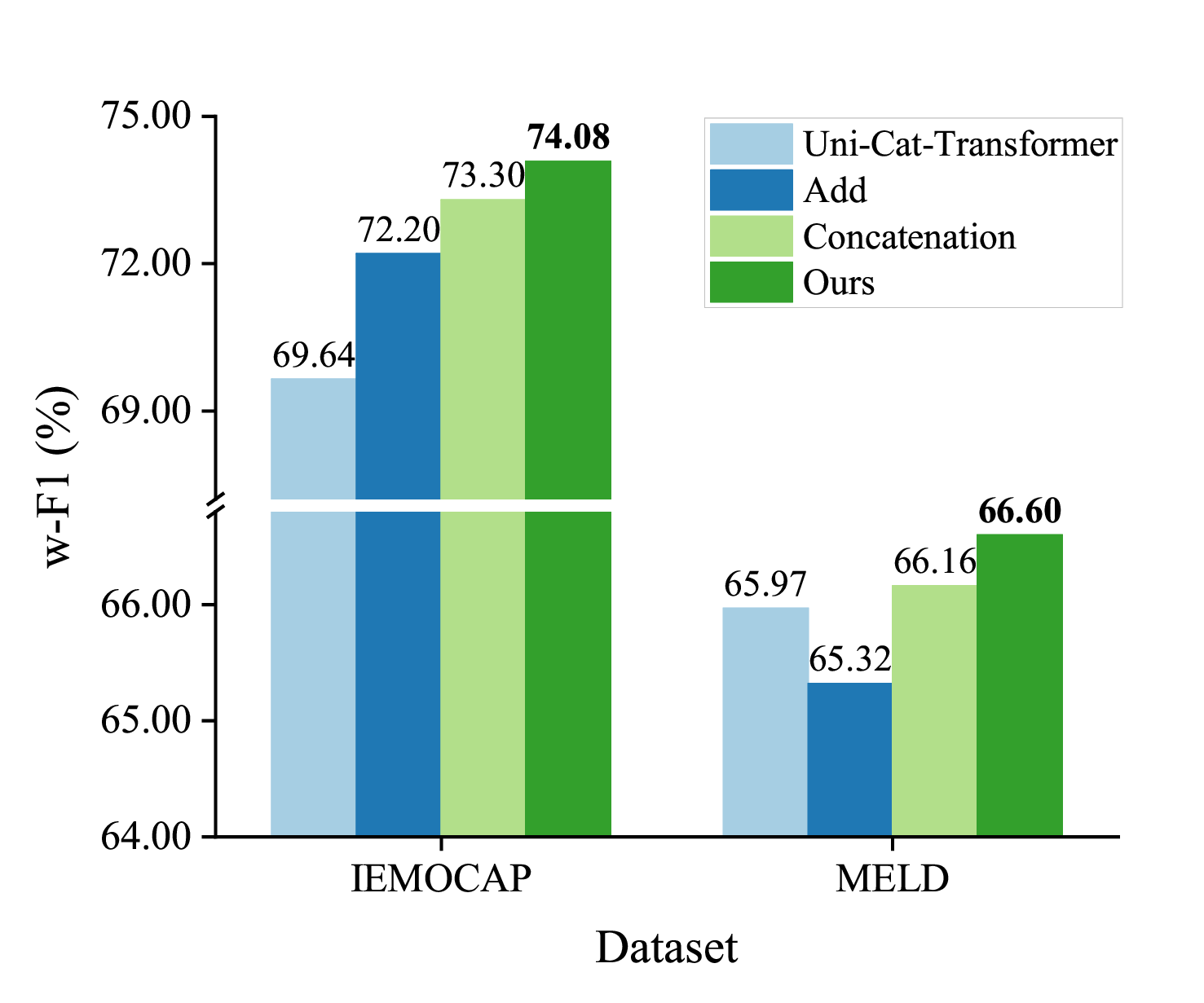}}
\hspace{-5mm}
\caption{Performance of different fusion methods on the two datasets. Bold font means that the improvement to all baselines is statistically significant (t-test with $p < 0.05$).}
\label{fig:3}
\end{figure}

To investigate the effect of our proposed hierarchical gated fusion module, we compare it with two typical information fusion strategies: (1) Add: representations are fused via element-wise addition. (2) Concatenation: representations are directly concatenated and followed by an FC layer. Add treats all representations equally, while Concatenation could implicitly choose the important information due to the FC layer. For a fair comparison, we replace the hierarchical gated fusion module of our model with hierarchical add and concatenation operations to implement the Add and Concatenation fusion strategies, respectively. In addition, we also compare SDT with a general transformer-based fusion method (i.e., unimodal features are concatenated and then fed into a transformer encoder) that we call Uni-Cat-Transformer.

As shown in Fig.~\ref{fig:3}, compared with other fusion strategies, our proposed hierarchical gated fusion strategy significantly outperforms them. The result indicates that directly fusing representations with Add and Concatenation is sub-optimal. Our proposed hierarchical gated fusion module first filters out irrelevant information at the unimodal level and then dynamically learns weights between different modalities at the multimodal level, which can more effectively fuse multimodal representations. 

In addition, our model achieves a significant performance improvement over Uni-Cat-Transformer that demonstrates the effectiveness of the proposed SDT in multimodal fusion. Interestingly, Uni-Cat-Transformer has poorer performance than Add and Concatenation on IEMOCAP; however, it shows an acceptable performance on MELD. This may be because interactions between modalities are not as complex on MELD as on IEMOCAP, and hence modeling modal interactions by multiple transformer encoders could generate some noise on MELD that makes Ui-Cat-Transformer gain comparable performance with Add and Concatenation. In contrast, SDT has superior performance than all baselines as it contains a hierarchical gated fusion module to filter out noise information, which further illustrates the usefulness of our proposed hierarchical gated fusion strategy. 

\subsection{Trends of Losses}
\begin{figure}[!t]
\hspace{-5mm}
\subfloat[${{\cal L}_{CE}}\,\&\,{{\cal L}_{Task}}$]{\label{fig:4a}
  \includegraphics[width=1.988in]{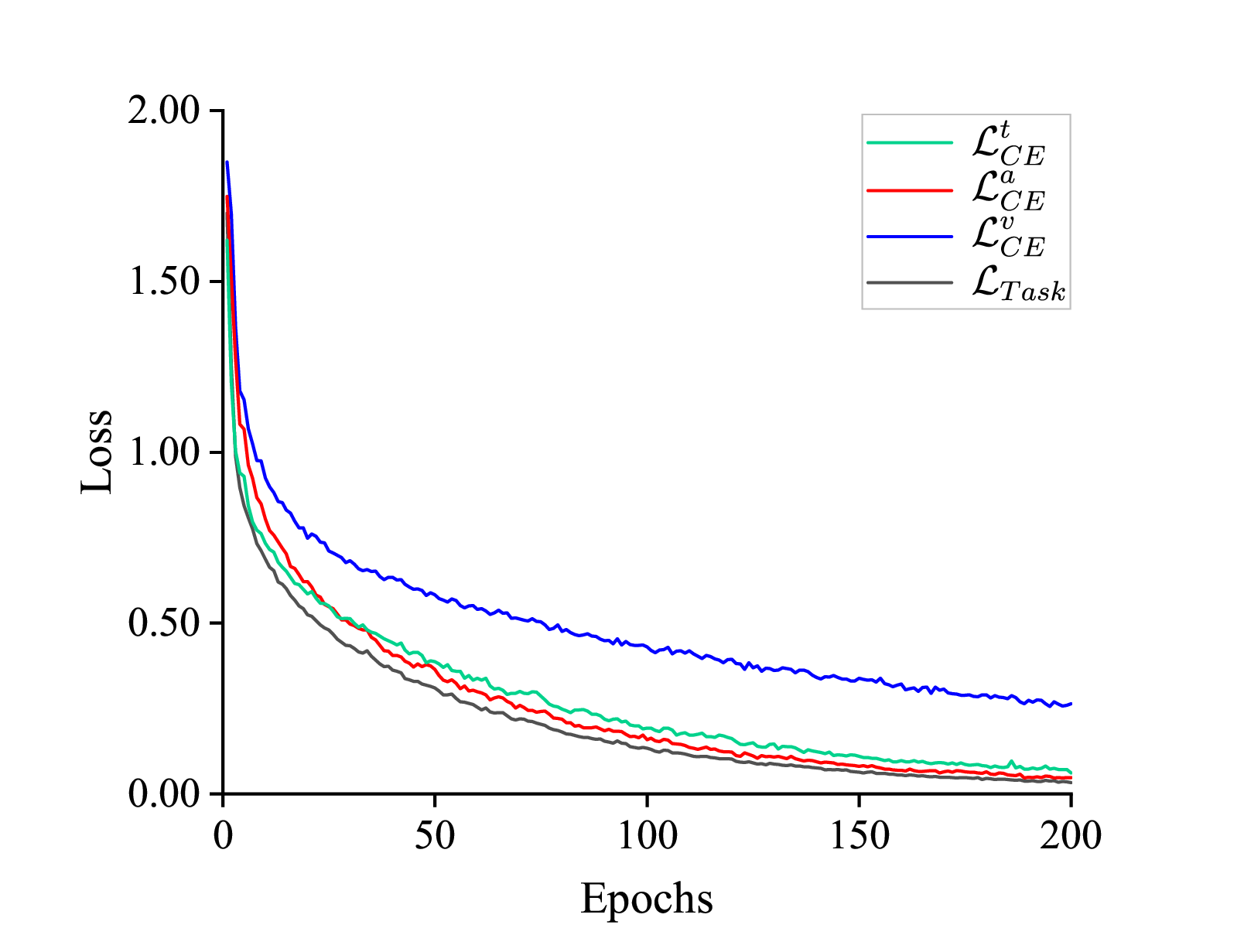}}
\hspace{-11.5mm}
\subfloat[${{\cal L}_{KL}}$]{\label{fig:4b}
  \includegraphics[width=1.988in]{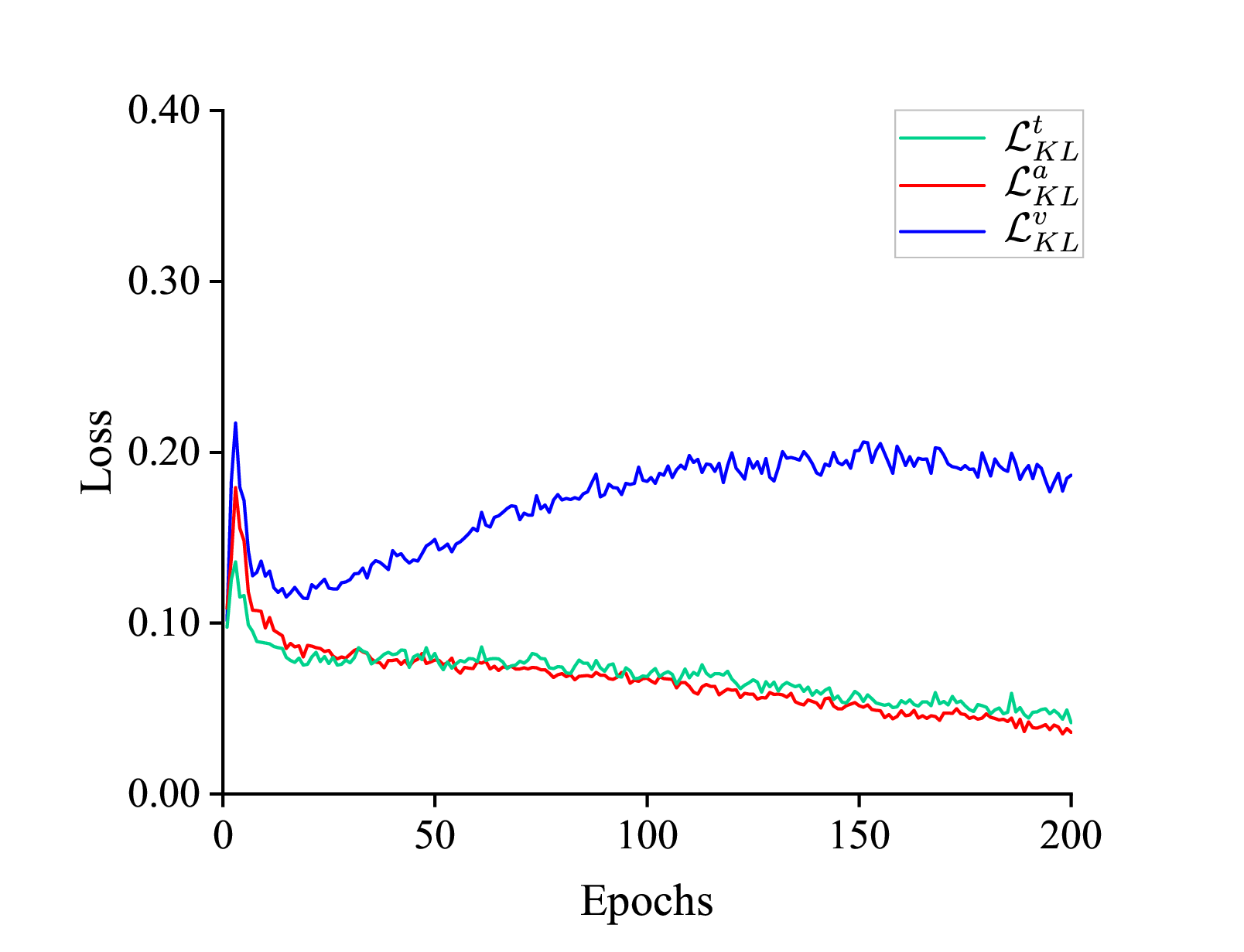}}
\caption{Trends of all losses during training on the IEMOCAP dataset.}
\label{fig:4}
\end{figure}

During training, we illustrate the trends of all types of losses on IEMOCAP dataset to better understand how these losses work, and Fig.~\ref{fig:4} displays the results. 

From Fig.~\ref{fig:4}\subref{fig:4a}, we find that ${\mathcal{L}}_{Task}$, ${\mathcal{L}}_{CE}^t$, ${\mathcal{L}}_{CE}^a$, and ${\mathcal{L}}_{CE}^v$ keep descending in the whole training process. From Fig.~\ref{fig:4}\subref{fig:4b}, we can see ${\mathcal{L}}_{KL}^t$ and ${\mathcal{L}}_{KL}^a$ also have decreasing trends except for fluctuations at the beginning, and ${\mathcal{L}}_{KL}^v$ goes down during early training except for the fluctuation and then goes up and achieves stability. Therefore, all of the losses can converge. These show that all students can learn knowledge from hard and soft labels to improve the model performance. Besides, we find that losses of student $v$ (i.e., ${\mathcal{L}}_{CE}^v$ and ${\mathcal{L}}_{KL}^v$) are larger than the other two students. This may due to a unsuitable learning rate for the student $v$. Hence, we would like to adaptively modify learning rates between different modalities to effectively optimize the proposed model in the future.

\subsection{Multimodal Representation Visualization}
\begin{figure*}
	\centering
	\subfloat[Origin representations]{\label{fig:5a}
		\includegraphics[width=1.850in]{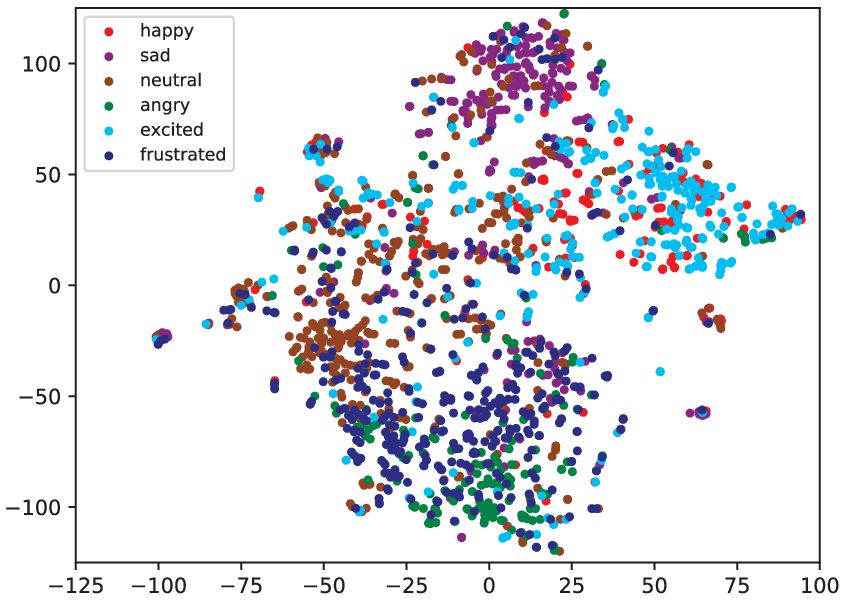}}
	\hspace{0.3in}
	\subfloat[Without self-distillation]{\label{fig:5b}
		\includegraphics[width=1.850in]{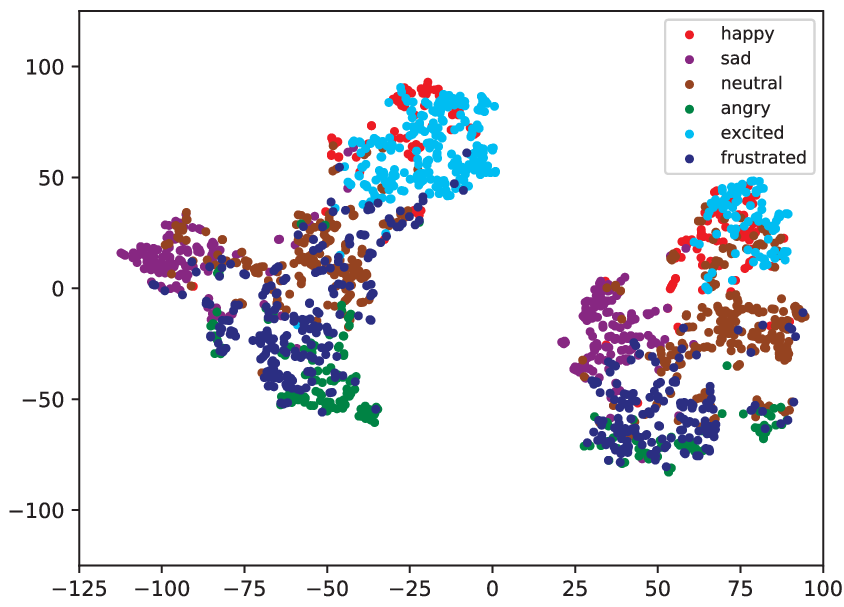}}
	\hspace{0.3in}
	\subfloat[With self-distillation]{\label{fig:5c}
		\includegraphics[width=1.850in]{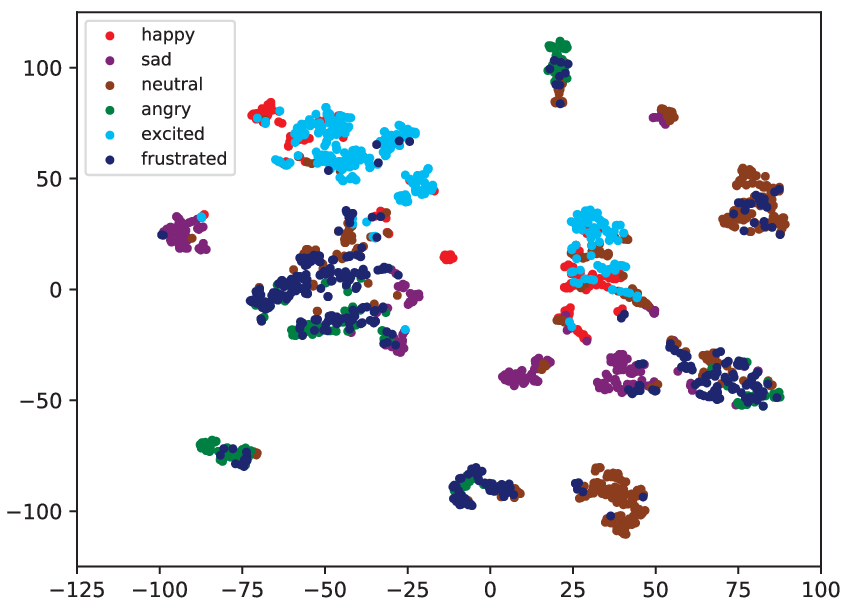}}
	\caption{t-SNE visualization of the multimodal representations with different emotion categories on the IEMOCAP dataset.}
	\label{fig:5}
\end{figure*}

\begin{figure*}
	\centering
	\subfloat[Origin representations]{\label{fig:5gendera}
		\includegraphics[width=1.850in]{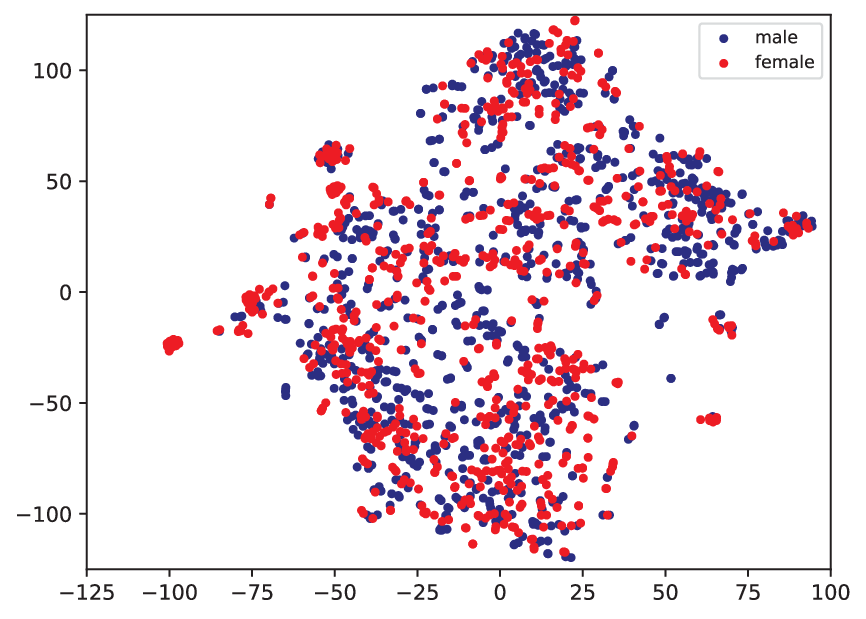}}
	\hspace{0.3in}
	\subfloat[Without self-distillation]{\label{fig:5genderb}
		\includegraphics[width=1.850in]{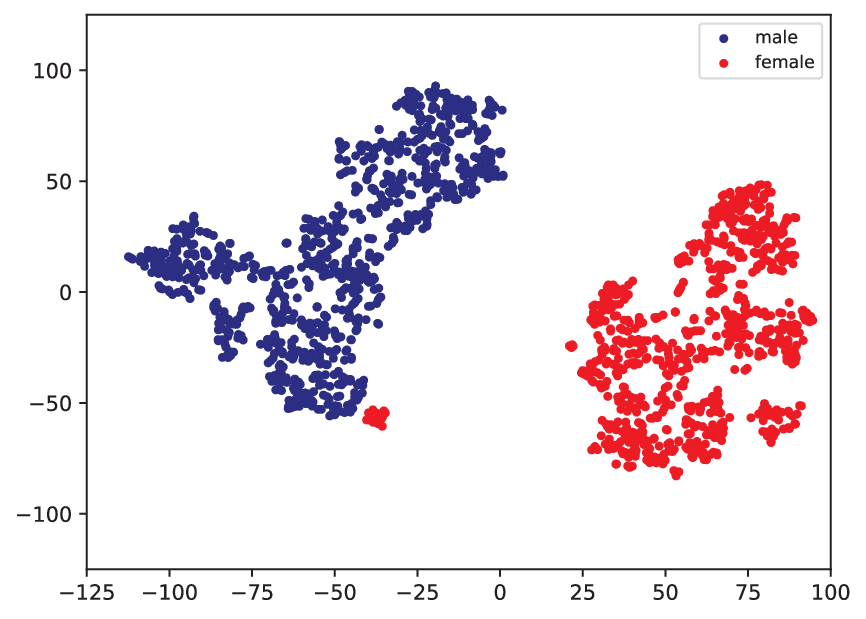}}
	\hspace{0.3in}
	\subfloat[With self-distillation]{\label{fig:5genderc}
		\includegraphics[width=1.850in]{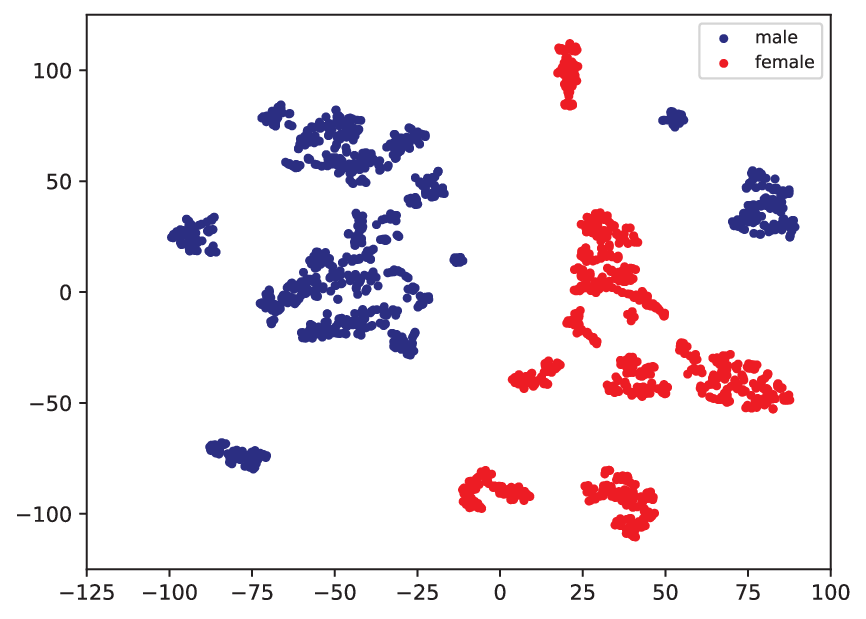}}
	\caption{t-SNE visualization of the multimodal representations  with different genders of speakers on the IEMOCAP dataset.}
	\label{fig:5gender}
\end{figure*}

We extract multimodal representations for each utterance on IEMOCAP from our proposed transformer-based model without and with self-distillation. Besides, pre-extracted unimodal representations are concatenated to produce original multimodal representations. Then, these multimodal representations are projected into two dimensions via the t-SNE algorithm \cite{van2008visualizing}.

Fig.~\ref{fig:5} illustrates the visualization results with different emotion categories. Compared with original multimodal representations, representations learned by the proposed transformer-based model become more clustered even without self-distillation. However without self-distillation,  multimodal representations of similar emotions (i.e., “happy” and “excited”, “angry” and “frustrated”) are difficult to separate; furthermore, representations of “neutral” emotion are intermingled with other emotions. By comparing Fig.~\ref{fig:5}\subref{fig:5b} and Fig.~\ref{fig:5}\subref{fig:5c}, we observe that our model with self-distillation yields a better separation and representations of different emotions are less mixed together. Therefore, introducing self-distillation training could learn more effective multimodal representations. 

On the other hand, we also show the visualization results with different genders of speakers in Fig.~\ref{fig:5gender}. Fig.~\ref{fig:5gender}\subref{fig:5genderb} and Fig.~\ref{fig:5gender}\subref{fig:5genderc} form two large clusters respectively corresponding to the gender of the speaker. This interesting finding indicates that with or without self-distillation, our model can distinguish the gender of the speaker, which may also be helpful for ERC.

\subsection{Case Study}
\begin{figure}[!t]
	\centering
	\includegraphics[width=3.5in]{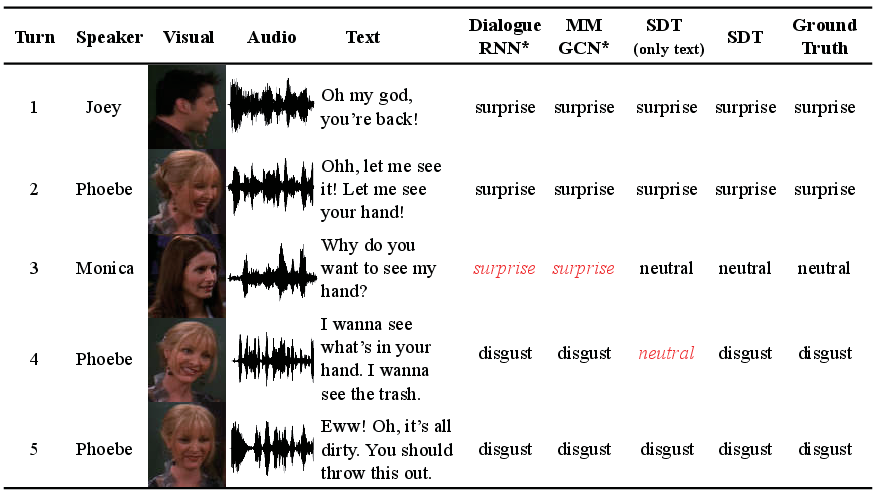}
	\caption{An example of emotion recognition results in a conversation from the MELD dataset.}
	\label{fig:6}
\end{figure}

To demonstrate the efficacy of SDT, we present a case study. Fig.~\ref{fig:6} shows a conversation that comes from MELD. SDT identifies the emotions of all utterances successfully, while DialogueRNN* and MMGCN* predict the 3rd utterance as ``surprise" incorrectly, probably because a question mark ``?" generally expresses ``surprise". This could indicate the more powerful multimodal fusion capability of our proposed model. On the other hand, using only the textual modality, our model recognizes the $4$th utterance as ``neutral" wrongly. To explore the reason behind it, we visualize multi-head attention weights of SDT (only text) and SDT for the $4$th utterance, respectively. For SDT, we find that the weights of textual features are obviously larger than acoustic and visual features for the $4$th utterance by outputting their weights. Therefore, we visualize only attention weights of the transformers that form enhanced textual modality representation in Fig.~\ref{fig:7}, and other visualization results can be found in the appendix.

As can be seen from Fig.~\ref{fig:7}\subref{fig:7a}, the $4$th utterance depends heavily on the $3$rd and $5$th utterances when using only the textual modality. The $3$rd utterance, which expresses ``neutral" emotion, may be more important due to a larger number of darkest attention heads; hence the $4$th utterance is identified as the same emotion as the $3$rd utterance, i.e., ``neutral". The utterance can be correctly recognized as ``disgust" by SDT for the following reasons: (1) According to Fig.~\ref{fig:7}\subref{fig:7b}, the text of the $4$th utterance is influenced the most by the text of the $5$th utterance whose emotion is ``disgust". (2) From Fig.~\ref{fig:7}\subref{fig:7c} and Fig.~\ref{fig:7}\subref{fig:7d}, we observe that the acoustic and visual expressions of the $2$nd, $4$th and $5$th utterances are more valuable for the $4$th utterance's textual expression, and the $4$th and $5$th utterances express ``disgust" emotion. Overall, the results show that interactions between modalities are helpful in identifying emotion from different perspectives, and therefore it is necessary to use multimodal information. In addition, comparing Fig.~\ref{fig:7}\subref{fig:7a} and Fig.~\ref{fig:7}\subref{fig:7b}, SDT learns better to correlate the $4$th utterance with the $5$th utterance. The finding illustrates that introducing self-distillation can learn more appropriate attention weights.

\begin{figure}
	\centering
	\subfloat[Intra-modal transformer \bm{$(t \to t)$} of SDT (only text)]{\label{fig:7a}
		\includegraphics[width=1.65in]{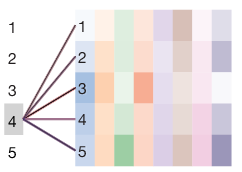}}
	\hspace{0.035in}
	\subfloat[Intra-modal transformer \bm{$(t \to t)$} of SDT]{\label{fig:7b}
		\includegraphics[width=1.65in]{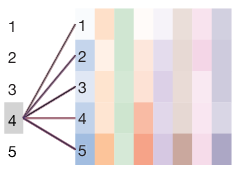}}
	\hspace{0.1in}
	\vspace{0.1in}
	\subfloat[Inter-modal transformer \bm{$(a \to t)$} of SDT]{\label{fig:7c}
		\includegraphics[width=1.65in]{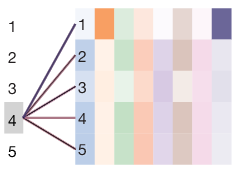}}
	\hspace{0.035in}
	\subfloat[Inter-modal transformer \bm{$(v \to t)$} of SDT]{\label{fig:7d}
		\includegraphics[width=1.65in]{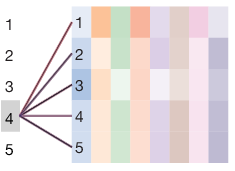}}
	\caption{Multi-head attention visualization for the $4$th utterance in Fig.~\ref{fig:6}. There are 8 attention heads and different colors represent different heads. The darker the color, the more important for the $4$th utterance.}
	\label{fig:7}
\end{figure}

\subsection{Error Analysis}
\begin{figure}[!t] 
\hspace{-3mm}
\subfloat[IEMOCAP]{\label{fig:8a}
  \includegraphics[width=1.65in]{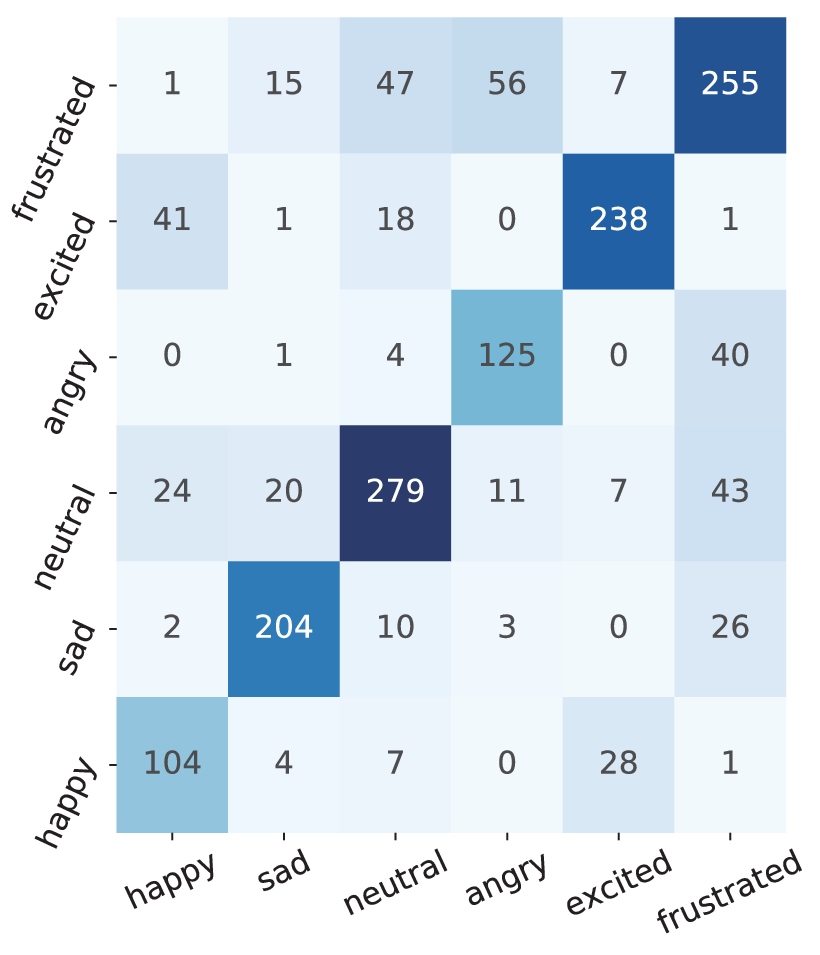}}
\subfloat[MELD]{\label{fig:8b}
  \includegraphics[width=1.75in]{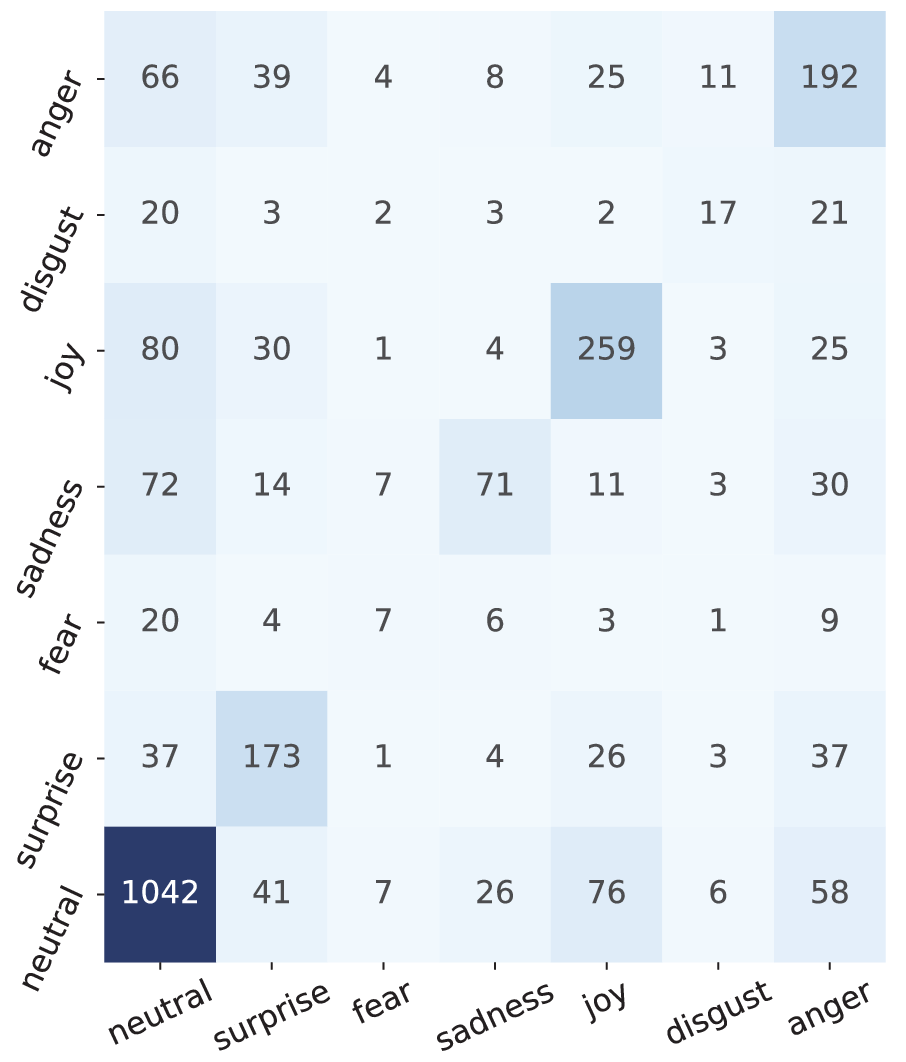}}
\hspace{-10mm}
\caption{The confusion matrices of the test set on the two datasets. The rows and columns represent true and predicted labels, respectively.}
\label{fig:8}
\end{figure}

Although the proposed SDT achieves strong performance, it still fails to detect some emotions. We analyze confusion matrices of the test set on the two datasets. From Fig.~\ref{fig:8}, we see that: (1) SDT misclassifies similar emotions, like ``happy" and ``excited", ``angry" and ``frustrated" on IEMOCAP, and ``surprise" and ``anger" on MELD. (2) SDT also tends to misclassify other emotions as ``neutral" on MELD due to that ``neutral" is the majority class. (3) It is difficult to correctly detect ``fear" and ``disgust" emotions on MELD because the two emotions are minority classes. Thus, it is challenging to recognize similar emotions and emotions with unbalanced data. 

\begin{table}
	\centering
	\caption{Test accuracy of SDT on utterances with and without emotional shift}
	\begin{tabular}{ccccc}
		\toprule
		\multirow{2}{*}{Dataset} & \multicolumn{2}{c}{Emotional Shift} & \multicolumn{2}{c}{w/o Emotional Shift} \\
		\cmidrule(lr){2-3}\cmidrule(lr){4-5} & 
		\multicolumn{1}{c}{\#Utterances} & \multicolumn{1}{c}{ACC} & \multicolumn{1}{c}{\#Utterances} & \multicolumn{1}{c}{ACC}  \\ \midrule
		IEMOCAP & 410 & 54.88 & 1151 &  80.71 \\
		MELD & 1003 & 61.62 & 861 & 73.05 \\ 
		\bottomrule
	\end{tabular}
	\label{tab:5}
\end{table}

Besides, we also investigate SDT performance on emotional shift (i.e., two consecutive utterances spoken by the same speaker have different emotions). As shown in Table~\ref{tab:5}, we observe that SDT performs poorer on utterances with emotional shift than that without it\footnote{In this paper, without emotional shift means two consecutive utterances spoken by the same speaker have same emotions.}, which is consistent with previous works. The emotional shift in conversations is a complex phenomenon caused by multiple latent variables, e.g., the speaker's personality and intent; however, SDT and most existing models do not consider these factors, which may result in poor performance. Further improvement on the case needs to be explored.

\section{Conclusion}
\label{sec:6}
In this paper, we propose SDT, a transformer-based model with self-distillation for multimodal ERC. We use intra- and inter-modal transformers to model intra- and inter-modal interactions between conversation utterances. To dynamically learn weights between different modalities, we design a hierarchical gated fusion strategy. Positional and speaker embeddings are also leveraged as additional inputs to capture contextual and speaker information. In addition, we devise self-distillation during training to transfer knowledge of hard and soft labels within the model to learn better modal representations, which could further improve performance. We conduct experiments on two benchmark datasets and the results demonstrate the effectiveness and superiority of SDT.

Through error analysis, we find that distinguishing similar emotions, detecting emotions with unbalanced data, and emotional shift are key challenges for ERC that are worth further exploration in future work. Furthermore, transformer-based fusion methods cause high computational costs as the self-attention mechanism of transformer has a complexity of $O\left( {{N^2}} \right)$ with respect to sequence length $N$. To alleviate the issue, Ding et al. \cite{ding2021sparse} proposed sparse fusion for multimodal transformers. Similarly, we plan to design a novel multimodal fusion method for transformers to reduce computational costs in the future.

\section*{Acknowledgments}
This work is partially supported by the Natural Science Foundation of China (No. 62006034), the Natural Science Foundation of Liaoning Province (No. 2021-BS-067), and the Fundamental Research Funds for the Central Universities (No. DUT21RC(3)015).

{\appendix[Attention Visualization]
	Multi-head attention weights of the transformers in our SDT that form enhanced acoustic and visual modality representations are visualized in Fig.~\ref{fig:9}.
	
\begin{figure}
	\centering
	\subfloat[Intra-modal transformer \bm{$(a \to a)$} of SDT]{\label{fig:9a}
		\includegraphics[width=1.65in]{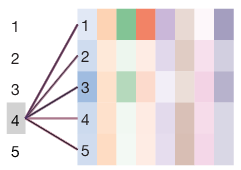}}
	\hspace{0.035in}
	\subfloat[Inter-modal transformer \bm{$(t \to a)$} of SDT]{\label{fig:9b}
		\includegraphics[width=1.65in]{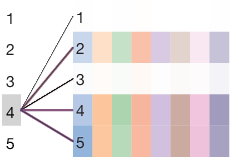}}
	\hspace{0.1in}
	\vspace{0.1in}
	\subfloat[Inter-modal transformer \bm{$(v \to a)$} of SDT]{\label{fig:9c}
		\includegraphics[width=1.65in]{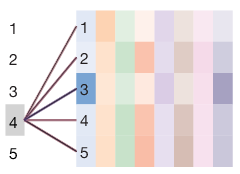}}
	\hspace{0.035in}
	\subfloat[Intra-modal transformer \bm{$(v \to v)$} of SDT]{\label{fig:9d}
		\includegraphics[width=1.65in]{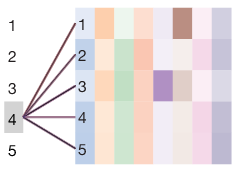}}
	\hspace{0.1in}
	\subfloat[Inter-modal transformer \bm{$(t \to v)$} of SDT]{\label{fig:9e}
		\includegraphics[width=1.65in]{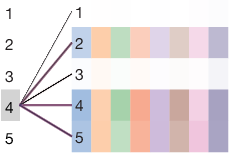}}
	\hspace{0.035in}
	\subfloat[Inter-modal transformer \bm{$(a \to v)$} of SDT]{\label{fig:9f}
		\includegraphics[width=1.65in]{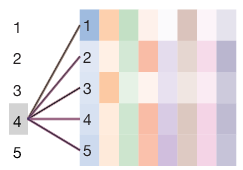}}
	\caption{Multi-head attention visualization for the $4$th utterance in Fig.~\ref{fig:6}.}
	\label{fig:9}
\end{figure}	
}

 
 \bibliographystyle{IEEEtran}
\bibliography{myreference.bib}

\newpage

\vspace{-33pt}
 \begin{IEEEbiography}[{\includegraphics[width=1in,height=1.25in,clip,keepaspectratio]{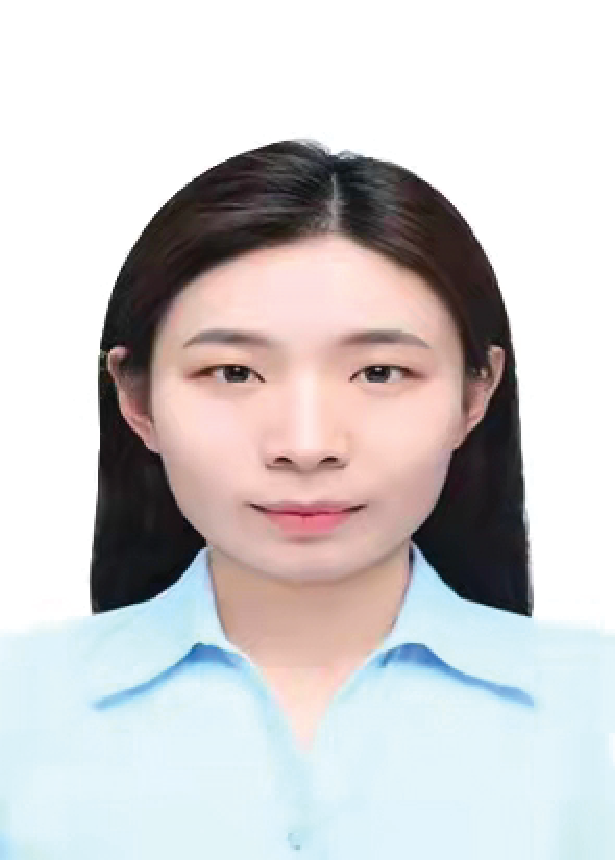}}]{Hui Ma}
received the M.S. degree from Dalian University of Technology, China, in 2019. She is currently working toward the Ph.D. degree at School of Computer Science and Technology, Dalian University of Technology. Her research interests include natural language processing, dialogue system, and sentiment analysis.
 \end{IEEEbiography}
\vspace{-33pt}
 \begin{IEEEbiography}[{\includegraphics[width=1in,height=1.25in,clip,keepaspectratio]{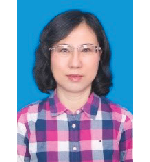}}]{Jian Wang}
	received the Ph.D. degree from Dalian University of Technology, China, in 2014. She is currently a professor at School of Computer Science and Technology, Dalian University of Technology. Her research interests include natural language processing, text mining, and information retrieval.
\end{IEEEbiography}
\vspace{-33pt}
\begin{IEEEbiography}[{\includegraphics[width=1in,height=1.25in,clip,keepaspectratio]{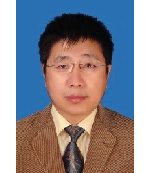}}]{Hongfei Lin}
	received the Ph.D. degree from Northeastern University, China, in 2000. He is currently a professor at School of Computer Science and Technology, Dalian University of Technology. His research interests include natural language processing, text mining, and sentimental analysis.
\end{IEEEbiography}
\vspace{-33pt}
\begin{IEEEbiography}[{\includegraphics[width=1in,height=1.25in,clip,keepaspectratio]{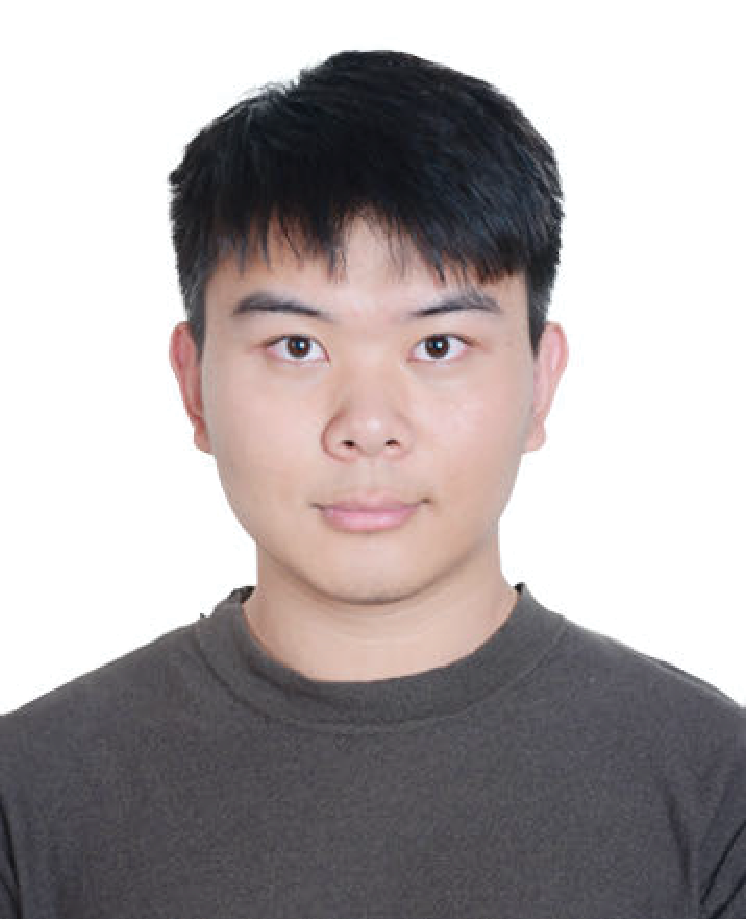}}]{Bo Zhang}
	received the B.S. degree from Tiangong University, China, in 2019. He is currently working toward the Ph.D. degree at School of Computer Science and Technology, Dalian University of Technology. His research interests include natural language processing, dialogue system, and text generation.
\end{IEEEbiography}
\vspace{-33pt}
\begin{IEEEbiography}[{\includegraphics[width=1in,height=1.25in,clip,keepaspectratio]{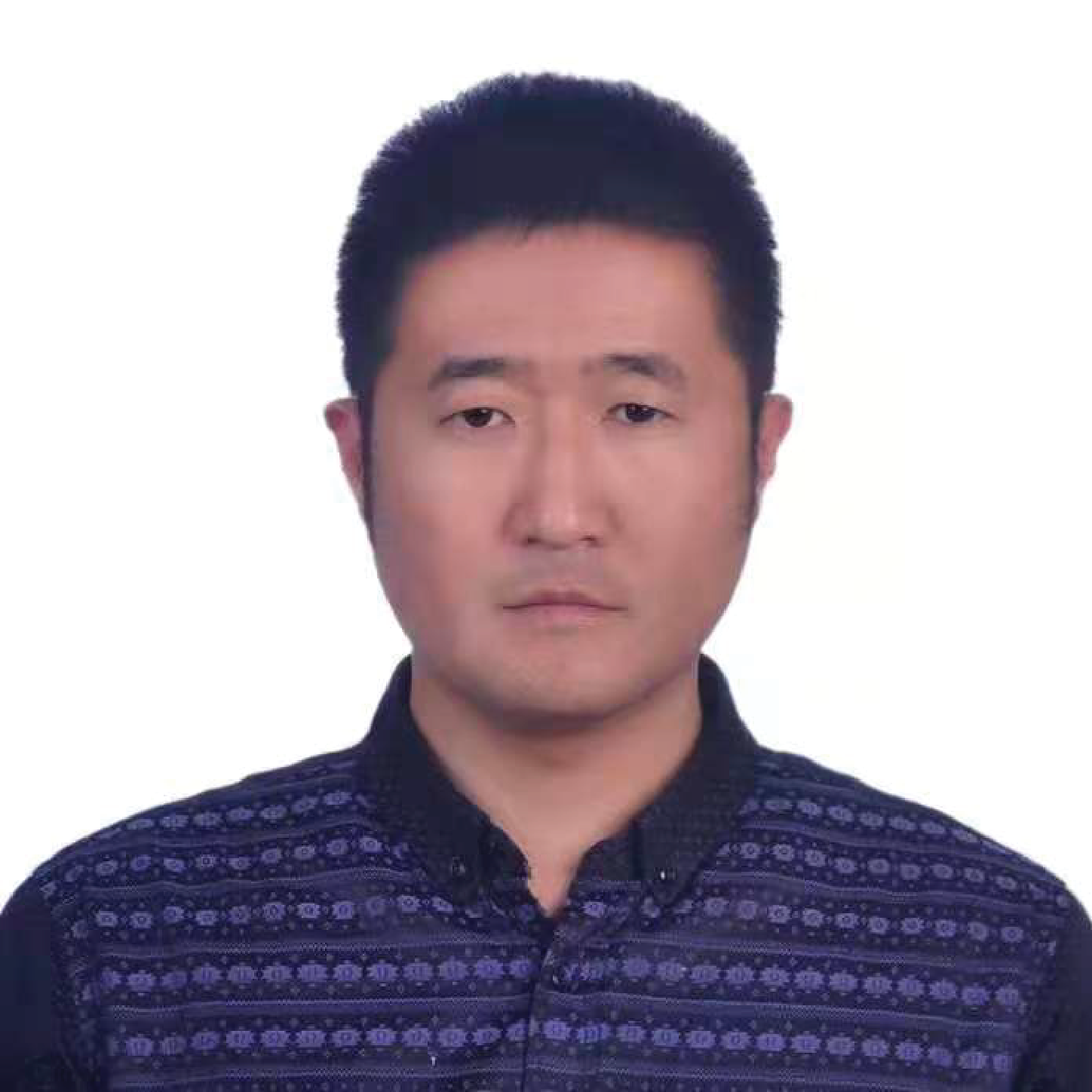}}]{Yijia Zhang}
	received the Ph.D. degree from the Dalian University of Technology, China, in 2014. He is currently a professor at School of Information Science and Technology, Dalian Maritime University. His research interests include natural language processing, bioinformatics, and text mining.
\end{IEEEbiography}
\vspace{-33pt}
\begin{IEEEbiography}[{\includegraphics[width=1in,height=1.25in,clip,keepaspectratio]{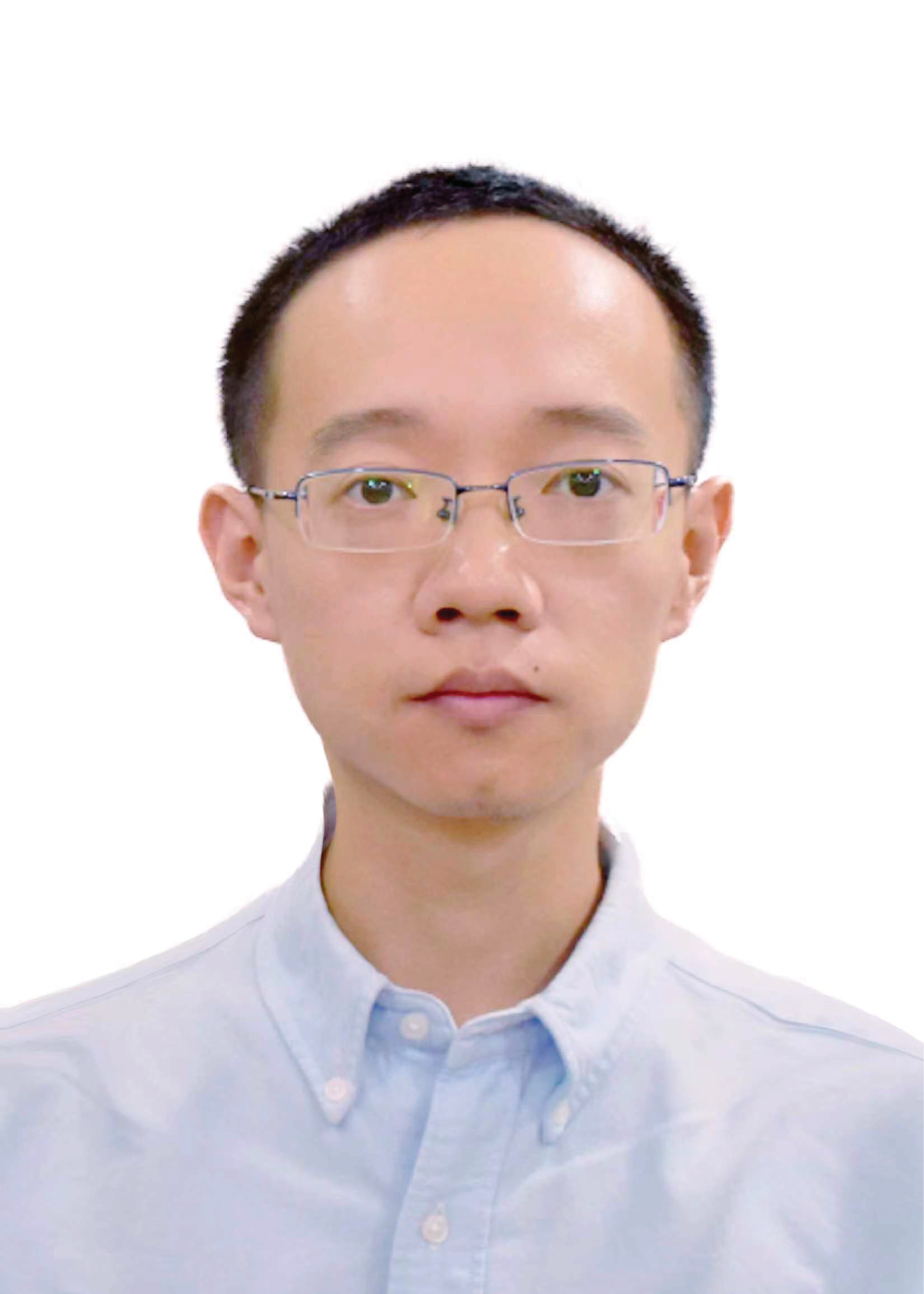}}]{Bo Xu}
	received the Ph.D. degree from the Dalian University of Technology, China, in 2018. He is currently an associate professor at School of Computer Science and Technology, Dalian University of Technology. His research interests include information retrieval, dialogue system, and natural language processing.
\end{IEEEbiography}




\end{document}